\documentclass{cas-dc}

\usepackage[numbers]{natbib}
\usepackage{hanging}
\usepackage{caption}

\begin{document}
\let\WriteBookmarks\relax
\def\floatpagepagefraction{1}
\def\textpagefraction{.001}
\shorttitle{GAN's Case Studies}
\shortauthors{Shamsolmoali et~al.}

\title [mode = title]{Image Synthesis with Adversarial Networks: a Comprehensive Survey and Case Studies}                      
%\tnotemark[Information Fusion]

%\tnotetext[2]{The second title footnote which is a longer text matter}
%   to fill through the whole text width and overflow into
%   another line in the footnotes area of the first page.}

\author[1, 2] {Pourya Shamsolmoali}
%\cormark[1]
\ead{pshams55@gmail.com}
\address[1]{Institute of Image Processing and Pattern Recognition, Shanghai Jiao Tong University, Shanghai, China.}

\author[1] {Masoumeh Zareapoor}
\ead{mzarea222@gmail.com}

\author[2] {Eric Granger}
\ead{Eric.Granger@etsmtl.ca}
\address[2]{Laboratoire d'imagerie, de vision et d'intelligence artificielle, \'Ecole de technologie sup\'erieure, Montreal, Canada.}

\author[3]{Huiyu Zhou}
\ead{hz143@leicester.ac.uk}
\address[3]{School of Informatics, University of Leicester, UK.}

\author[4]{Ruili Wang}
\ead{Ruili.wang@massey.ac.nz}
\address[4]{School of Natural and Computational Sciences, Massey University, Auckland, New Zealand.}

\author[5]{M. Emre Celebi}
\ead{ecelebi@uca.edu}
\address[5]{Department of Computer Science, University of Central Arkansas, USA.}

\author[1] {Jie Yang}
%\cormark[1]
\ead{jieyang@sjtu.edu.cn}

\begin{abstract}
Generative Adversarial Networks (GANs) have been extremely successful in various application domains such as computer vision, medicine, and natural language processing. Moreover, transforming an object or person to a desired shape become a well-studied research in the GANs. GANs are powerful models for learning complex distributions to synthesize semantically meaningful samples. However, there is a lack of comprehensive review in this field, especially lack of a collection of GANs loss-variant, evaluation metrics, remedies for diverse image generation, and stable training. Given the current fast GANs development, in this survey, we provide a comprehensive review of adversarial models for image synthesis. We summarize the synthetic image generation methods, and discuss the categories including image-to-image translation, fusion image generation, label-to-image mapping, and text-to-image translation. We organize the literature based on their base models, developed ideas related to architectures, constraints, loss functions, evaluation metrics, and training datasets. We present milestones of adversarial models, review an extensive selection of previous works in various categories, and present insights on the development route from the model-based to data-driven methods. Further, we highlight a range of potential future research directions. One of the unique features of this review is that all software implementations of these GAN methods and datasets have been collected and made available in one place at {\fontfamily{qcr}\selectfont https://github.com/pshams55/GAN-Case-Study}. 
\end{abstract}

%\includegraphics{figs/grabs.pdf}

%\item Research highlights item 1
%\item Research highlights item 2
%\item Research highlights item 3

\begin{keywords}
GANs \sep Image Synthesis \sep Image-to-Image Translation \sep Image Fusion \sep Classification 
\end{keywords}

\maketitle

\section{Introduction}
\begin{figure}
\centering
\includegraphics[width=3.4in]{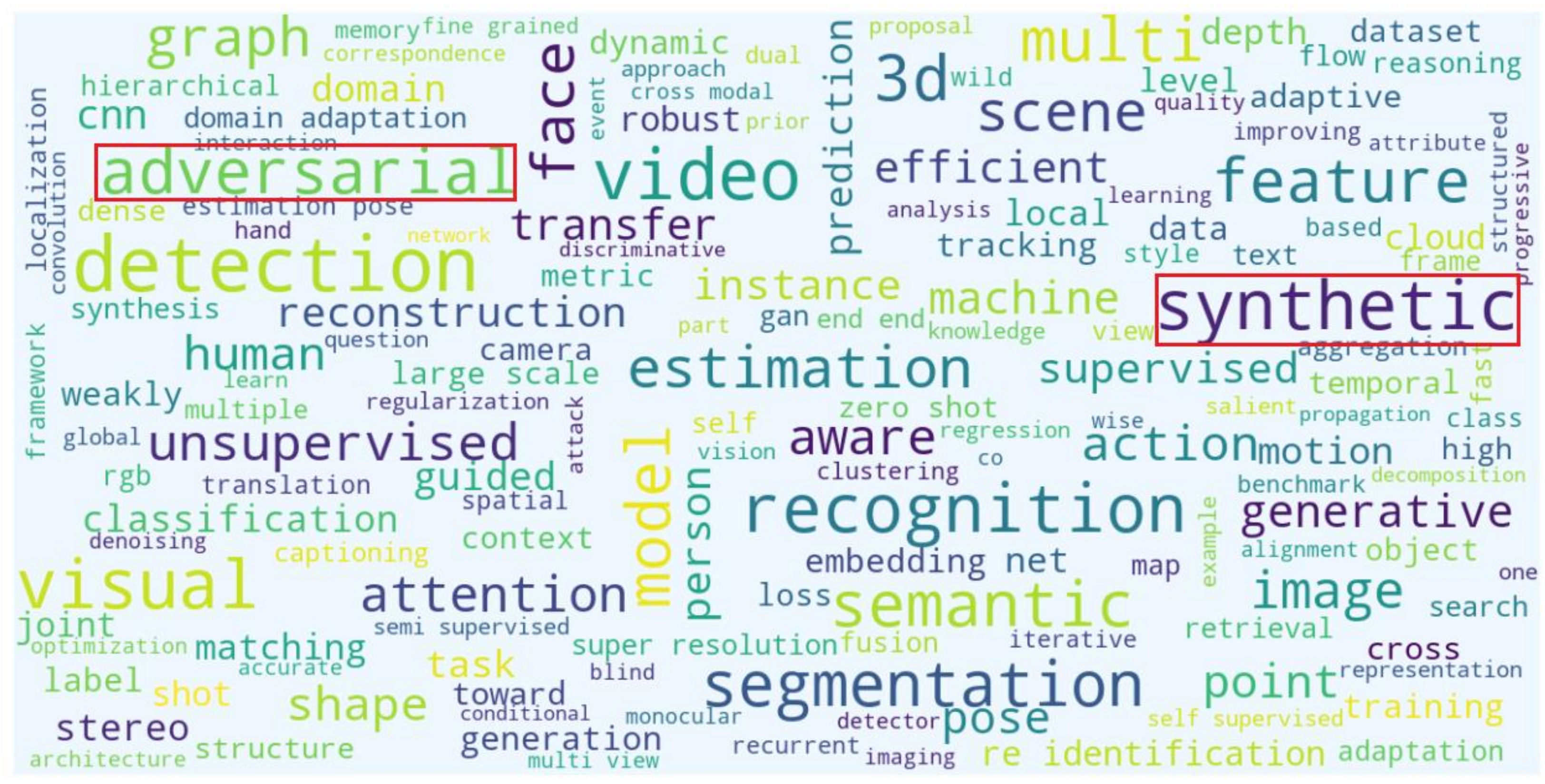}
\caption{It shows the most frequent keywords in CVPR and ICCV conferences from 2017 to 2020. The size of each word indicates the frequency of that keywords. The, {\it "synthetic" and "adversarial"} keywords are often have been used.}
\label{fig:s1b}
\end{figure}

Big data has enabled deep learning algorithms achieve rapid advancements. In particular, state-of-the-art generative adversarial networks (GANs) \cite{royer2020xgan} are able to generate high-fidelity natural images of diverse categories. It is demonstrated that, given proper training, GANs are able to synthesize semantically meaningful data from standard data distributions. 
The GAN was introduced by Goodfellow et al. \cite{goodfellow2014generative} in 2014, and performs better than other generative models in producing synthetic images, and later has become an active research area in computer vision. Figure \ref{fig:s1b} shows the importance of this topic in the recent years. The standard GAN contains two neural networks, a generator and a discriminator, in which the generator attempts to create realistic samples that deceive the discriminator, which strives to distinguish the real samples from the fake ones. The training procedure continues until the generator wins the adversarial game.Then, the discriminator makes the decision that a random sample either is fake or real.
There are two main research directions in GAN. The first is focused on the theoretical thread that attempts to improve GAN stability, and address the training issues of GAN \cite{radford2015unsupervised, salimans2016improved, hoang2018mgan, engelhardt2018improving, metz2016unrolled}, or reformulate it from different viewpoints like information theory \cite{chen2016infogan, song2019generative, olmos2019binocular} and efficiency \cite{zhao2016energy, balaji2019entropic, schafer2019implicit}. The second focuses on the architectures and applications of GAN in computer vision \cite{radford2015unsupervised, dumoulin2016guide, karras2019style}. In addition to image synthesis, there are numerous applications where GAN is successfully used, such as image super-resolution \cite{ledig2017photo}, image captioning \cite{zhao2017dual}, image inpainting \cite{wang2017shape}, text-to-image translation \cite{zhang2017stackgan}, semantic segmentation \cite{luc2016semantic}, object detection \cite{mokhayeri2020paired}, generative adversarial attack \cite{chakraborty2018adversarial}, neural machine translation \cite{yang2017improving}, image fusion \cite{ma2020pan, zhan2019spatial, joo2018generating, rey2020fucitnet} and image denozing \cite{park2019unpaired}. 
\\

An introduction to GANs has been provided by Huang et al. \cite{huang2018introduction} and Goodfellow \cite{goodfellow2016nips} where they discussed the significance of GAN models and compares GANs and its variations to generate synthetic samples. More recently, Creswell et al. \cite{creswell2018generative} presented a survey of GANs which evaluates models and training methods. These generic image synthesis surveys, discussed GANs in a general context, without considering the formation details, advantages and drawback of each model. These studies are limited in that they only cover GANs architectures and algorithmic methods including feature selection and weighted approaches. The GANs surveys that have been published so far are summarized in Table~\ref{table:tab11}. However, there is a lack of a survey paper that with theoretical analysis discusses the advantages and disadvantages of presented models.
\begin{figure}
 \vspace{-5mm}
 \hspace{-1cm}
  \includegraphics[width=3.5in]{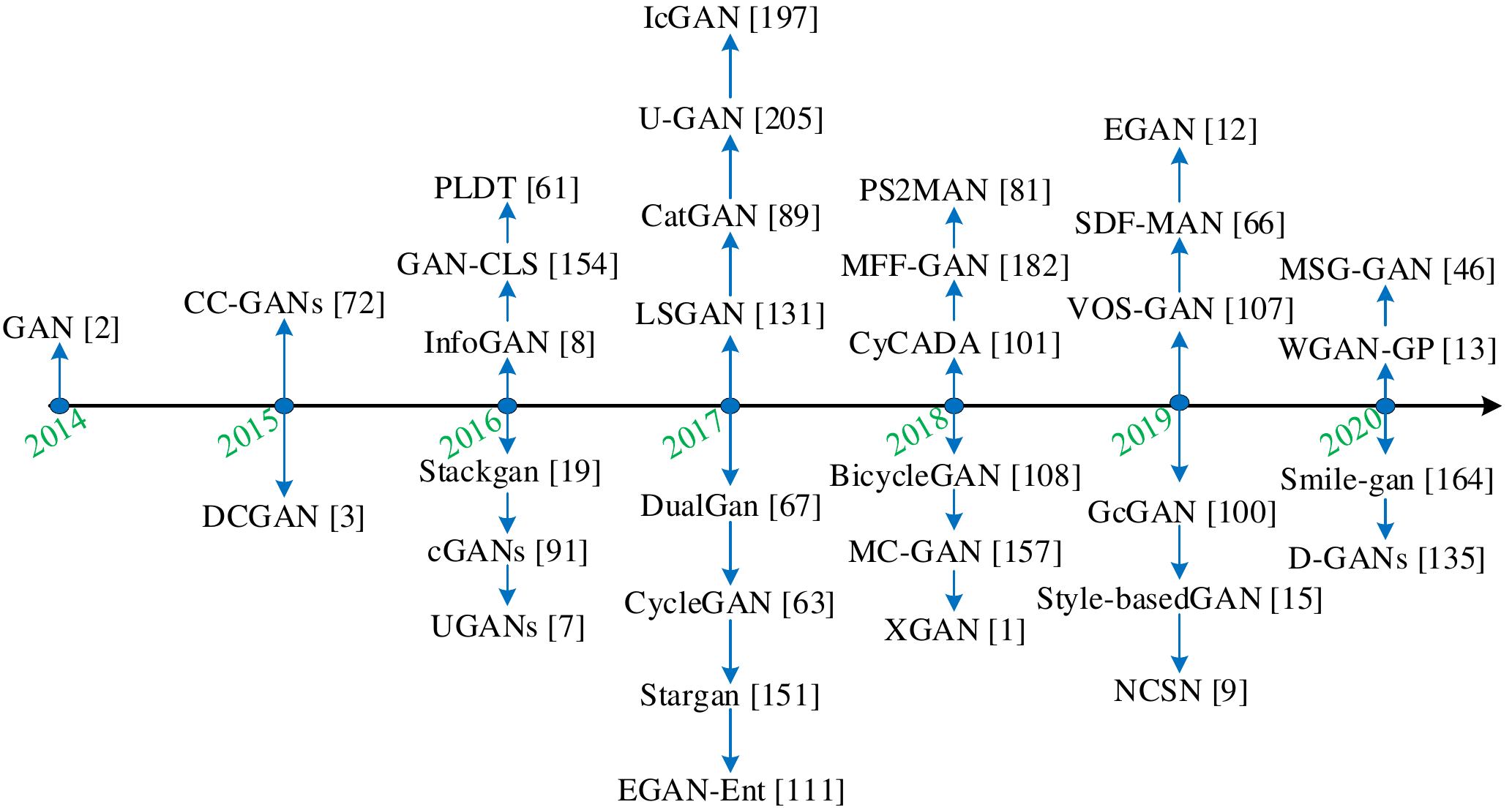}
%\centering
\caption{An overview of image synthesis methods since 2014, we tried to reference the original research articles, later discuss their proposed variations in the content.}
\label{fig:s1a}
\end{figure}
\\

Based on these observations, this paper targets the following open questions. What are the current state-of-the-art GANs for image synthesis? Is there currently any efficient GAN models that reach state-of-the-art performance for synthetic image generation? What type of GANs architectures works best for image-to-image translation? Which loss functions are most effective? And finally: we summarized the datasets that are widely used to validate different GANs based approaches. 
Milestones of GANs for generating synthesis images are listed in Figure~\ref{fig:s1a}.  
\\

In this paper, we provide an empirical comparative study of GAN models for synthetic image generation. We show how GAN can be trained efficiently to learn hidden discriminative features. For fair comparison between the tested approaches, we used a common framework in Python and Tensorflow to train the models with 4 NVIDIA GTX Geforce 1080 Ti GPUs. 
The outline of our review contains the definition of the problems, a summary of the main GAN methods, and the detailed coverage of the specific solutions. We reserve a detailed section for all the synthetic image generation benchmarks that are related to GANs. Moreover, this survey discusses the advantages and disadvantages of current GAN models with the mathematical foundations and theoretical analysis.
Finally, an accompanying web-page as a living repository of papers that address GANs for image synthesis problems is structured based on our taxonomy. This web-page will be continuously updated with new studies and approaches {\url{https://github.com/pshams55/GAN-Case-Study}}. 
The main contribution of this paper can be summarized as follows:
\begin{itemize}
\item A taxonomy of recent applications of GANs for synthetic image generation in various domains under two categories: single-stage and multi-stage models.
\item Review of the architecture of state-of-the-art models designed for GANs. 
\item We provide the details of performance metrics that are generally use to evaluation the GAN models and datasets.
\item We provide the community a live repository which contains the source codes, datasets and papers that are discussed in this survey. It also will be updated monthly. 
\end{itemize}

The rest of the paper is organized as follows. The next section reviews the core concepts of GANs. Section \ref{sec:3} discusses different loss functions that are used in GANs. Section \ref{sec:4} discusses the most commonly used datasets for evaluating GAN models. Section \ref{sec:5} presents the approaches that are used in image synthesis. Section \ref{sec:6}, evaluated the methods that are used for image-to-image synthesis, and possible future research areas. In Section \ref{sec:7}, we present a summary discussion and discuss limitations of GAN based methods.

%\begin{enumerate}
%\itemsep=0pt
%\item {natbib.sty} for citation processing;
%\item {geometry.sty} for margin settings;
%\item {fleqn.clo} for left aligned equations;
%\item {graphicx.sty} for graphics inclusion;
%\item {hyperref.sty} optional packages if hyperlinking is
%  required in the document;
%\end{enumerate}  
\begin{table*} 
\caption{Summary of GAN surveys for different applications since 2017.}
\label{table:tab11}
%\hspace{-15mm}
\scriptsize
%\resizebox{\textwidth}{!}{%
\begin{tabular}{p{.4\textwidth}p{.15\textwidth}p{.05\textwidth}p{.3\textwidth}}
\hline \\ [0.1ex]
Title &  Conference/Journal & Year & Short descriptions\\  
\hline\\ [0.3ex]
An Overview and Comparative Analysis on Major Generative Models & NeurIPS \cite{gu2018overview} & 2017 & Short survey statistically evaluates four important generative model: VAE, WGAN-GP, WINN, and DCGAN, over CIFAR-10 and CelebA \\
Generative Adversarial Networks: An Overview & IEEE SPM \cite{creswell2018generative} &  2017 & It provides an overview of GANs for the signal processing community, speech and natural language processing \\ 
A Large-Scale Study on Regularization and Normalization in GANs & JMLR \cite{kurach2019large} & 2018 & Discusses the impact of regularization and normalization schemes on GANs training; authors provide the important evaluation metrics in GANs methods \\ 
Efficient GAN-Based Anomaly Detection & ICDM \cite{zenati2018efficient} &  2018 & Short review of GAN-based anomaly detection methods \\ 
Generative Adversarial Network in Medical Imaging: A Review & MedIA \cite{yi2019generative} & 2019 &  A broad survey of the advanced methods in medical imaging using the adversarial training scheme with a comprehensive evaluation results  \\ 
Adversarial Training in Affective Computing and Sentiment Analysis: Recent Advances and Perspectives & IEEE CIM \cite{han2019adversarial} & 2019 & A brief review for GANs architectures in term of stability and efficiency\\ 
Stabilizing Generative Adversarial Network Training: A Survey & ICML \cite{wiatrak2019stabilizing} & 2019 & This survey provides a brief explanation towards stabilizing the GANs training process; categorizing the issues in the training GANs \\ 
Generative Adversarial Networks in Computer Vision: A Survey and Taxonomy & ArXiv \cite{wang2019generative} & 2019 &  This paper reviewed and critically discussed the most popular architecture-variant, and loss-variant GANs. \\
How generative adversarial networks and their variants work: An Overview  & ACM CSur \cite{hong2019generative} & 2019 & A brief review of GAN-based methods \\ 
A Survey on GANs for Anomaly Detection & ICML  \cite{di2019survey} & 2019 & Discusses the efficiency, structure and empirical validation of the main GAN models that are proposed for anomaly detection \\ 
A Survey on Deep Learning in Medicine: Why, How and When & INFFUS \cite{piccialli2020survey} & 2020 & A brief review focus on the deep learning and GANs applied in the medicine  \\
Single Image Deraining: From Model-Based to Data-Driven and Beyond & IEEE TPAMI \cite{yang2020single} & 2020 & A Comprehensive review of deraining methods over the last decade (using GANs \& CNNs) \\
An attention-based Unsupervised Adversarial Model for Movie Review Spam Detection & IEEE TMM \cite{gong2020attention} & 2020 &  A comprehensive review of GAN-based unsupervised spam detection with a thorough and detailed evaluation results \\ 
%How Generative Adversarial Networks and Their Variants Work: An Overview & ACM Computing Surveys \cite{hong2019generative} & 2019 &  This review focused on how a GAN can be combined with autoencoder frameworks \\
A Review on Generative Adversarial Networks: Algorithms, Theory, and Applications & ArXiv \cite{gui2020review} & 2020 &  This paper discussed the theoretical issues related to GAN models. \\
\hline
\end{tabular}
%}
\end{table*}

\section{Related Work}
\begin{figure*}
  \centering
  \includegraphics[width=160mm, height=50mm]{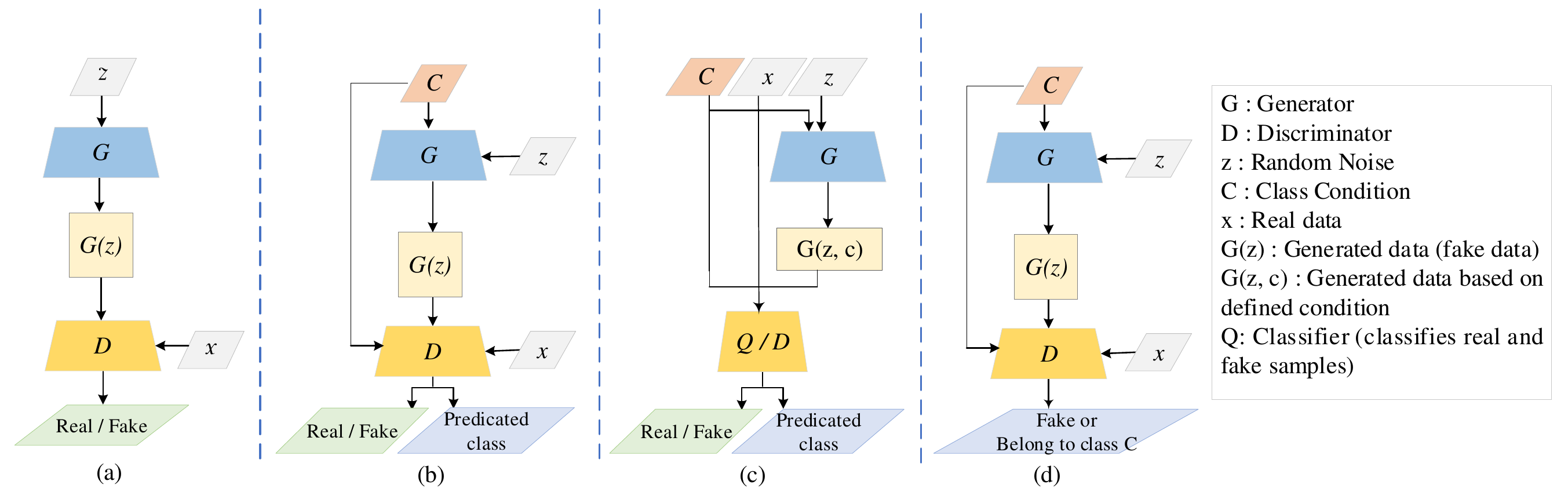}
  \caption{GAN architectures and its variations.{\bf (a)} Shows the basic architecture of a GAN \cite{goodfellow2014generative}; {\bf (b)}  Shows the conditional GAN, introduced in \cite{isola2017image} that perform class-conditional image synthesis; {\bf (c)} Shows the architecture of  InfoGAN \cite{chen2016infogan} and ACGAN \cite{odena2017conditional}; and {\bf (d)} Shows the BAGAN architecture \cite{mariani2018bagan}.}
\label{fig:s2}
\end{figure*}

\begin{figure}[b]
  \centering
  \includegraphics[width=3.2in]{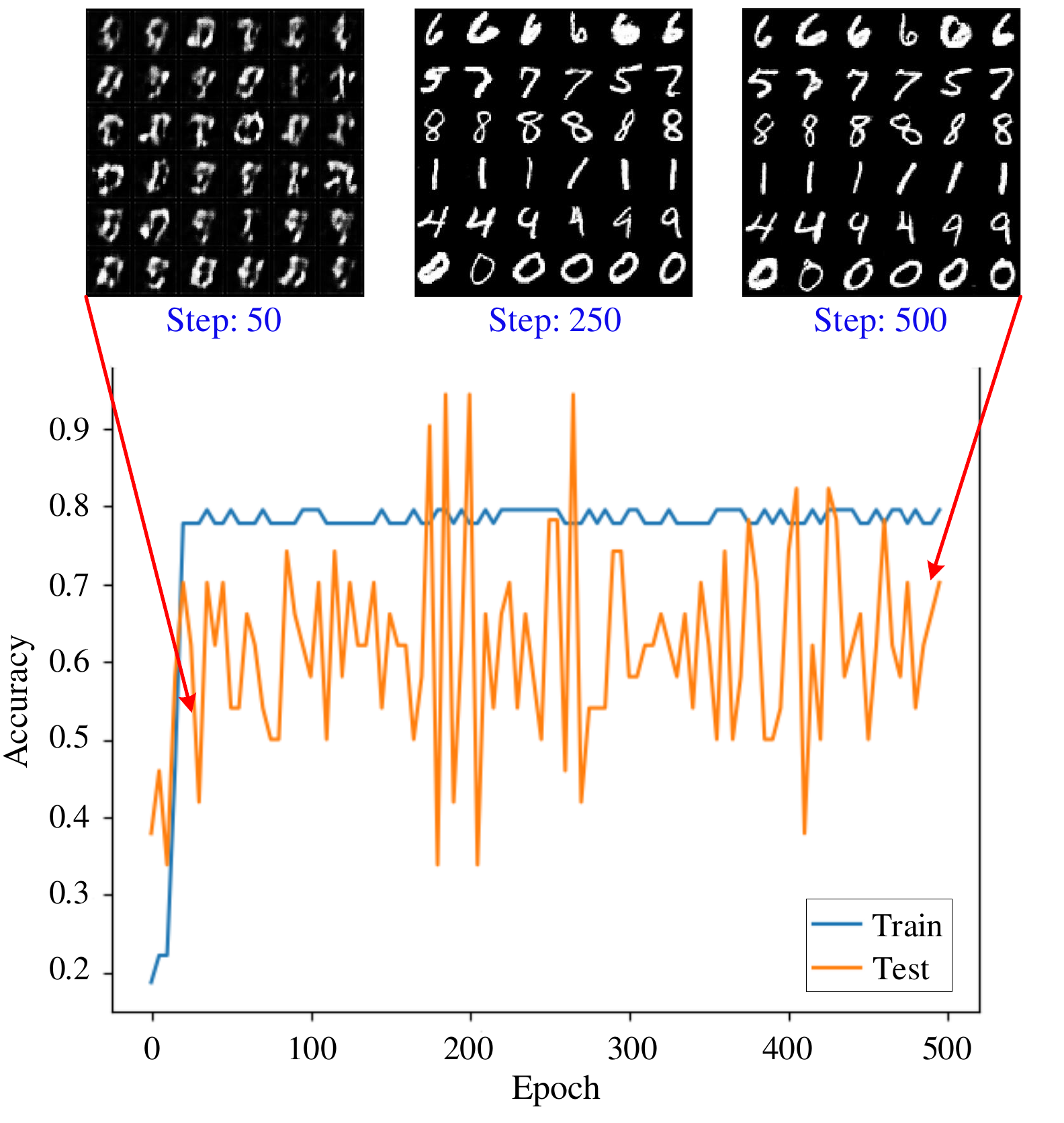}
  \caption{Throughout the training, the $G$ is tries to generate a distribution of samples which have similar distribution to real data. Generally, the GANs are represented by the learned parameters (weights) that have been captured from $G$ and $D$.}
\label{fig:s3}
\end{figure}

In this section, we first enlist some of the famous works that have benefited from GANs, then we focus on the various applications including medical imaging and image-to-image translation. The literature review shows few review papers on GAN architectures and performance are available \cite{creswell2018generative, wu2017survey}. Those works mostly focused on the performance validation for the different types of GANs architectures. The others works are limited because the benchmark datasets do not reflect the diversity in a proper way. Hence, the results mostly focus on image quality assessment, which may discount GANs effectiveness in generating diverse images \cite{karnewar2020msg}. In this paper, we gather a wide range of GAN models and discuss them in details. To avoid interruptions in the flow of our exposition, we first present the original GAN definition and then illustrate its variations in next subsections.  
A generative G parameterized by $\theta$ and receives random noise $\verb|z|$ as input and output will be sample  {\itshape G $(z,\theta)$}. Hence the output can be a sample generated from the distribution: {\itshape G $(z,\theta), p_g$}. Moreover, there are a massive training data {\itshape x} received from $p_{data}$, and the objective of the {\itshape G} is to approximate $p_{data}$ while using $p_g$. The basic architecture of GAN is presented in Figure~\ref{fig:s2}(a). 
GAN \cite{goodfellow2014generative} contains two different neural networks: a generator \textit{G} that as a input receives random noise vector  $\verb|z|$, and produces synthetic data $\verb|G(z)|$, the discriminator \textit{D} gets both the real data \verb|x| and the generated data $\verb|G(z)|$ as an input and distinguishes the real sample from the fake one, as presented in Figure~\ref{fig:s2}(a). In general, both \textit{G} and \textit{D} have neural network architectures. The first architecture of GAN \cite{goodfellow2014generative} uses fully connected layers $\verb|FC|$ as the base network. Later, Radford et al. \cite{radford2015unsupervised} introduced a deep fully convolutional neural networks in GAN which improves the results, and it was the start to largely use convolution layers in many GAN models. Additional details regarding convolution arithmetic can be found in \cite{dumoulin2016guide}. The main idea for training the GAN is to organize a two-player min-max game in which the \textit{G} attempts to produce accurate data for D which tries to recognize the real data from the fake ones \cite{goodfellow2014generative}. The value function is describes in Eq.(\ref{eq:e1}) where $p_{data}(x)$ signifies the real data distribution and $p_z(z)$ represents the noise distribution.
\begin{equation}
\begin{aligned}
  \min_{G}\max_{D}V(G,D)=E_{x\sim P_{data}(x)}[\log D(x)]+ \\
  E_{z\sim P_z (z)}[\log (1-D(G(z)))]
  \end{aligned}
\label{eq:e1}
\end{equation}
$G$, $D$ is a generator and discriminator, $G$ maps $z$ into the element $X$, and $D$ takes both the input from the input $x$ and $z$ to classify them as real or fake/ generated samples. If a sample comes from real input x, $D$ will maximize its output, while if a sample comes from $G$ (generated samples), $D$ will minimize its output, thus, the $2^{nd}$ term of the Eq.(\ref{eq:e1}) (log (1-D(G(z)))) will be appeared. This adjustment led to two aspects. {\fontfamily{qcr} \selectfont i)} The decision boundary made by $D$ disciplines enormous error to the created samples that have distance from the decision boundary, thereby helps the {\fontfamily{cmr} \selectfont "bad"} produced samples moving toward the decision boundary. This approach is helpful for generating high quality images; {\fontfamily{qcr} \selectfont ii)} The produced samples that are not close to decision boundary can deliver more gradient when updating the $G$, which solves the vanishing gradient problems of GAN training. Figure~\ref{fig:s3} shows the training steps of a standard GAN on MNIST dataset. It is worth mentioning that, there are three {\fontfamily{qcr} \selectfont key challenges} with the GANs \cite{wiatrak2019stabilizing}:
\begin{itemize}
\item {\it Mode collapse}: Concerns are not restricted to the process of reaching the equilibrium. One of the most frequent failures of GANs is {\it mode collapse}, which occurs while the G maps various diverse inputs to the same output.
\item {\it Vanishing gradients}: For optimal training of a GAN, both {\it G} and {\it D} need to generate valuable feedback. A well-trained {\it D} squashes the loss function to 0, consequently, gradients are approximately zero, which delivers a small number of feedbacks to the {\it G} that resulted in slowing or completely stopping the learning. Equally, an inaccurate {\it D} generates wrong feedback which misleads the {\it G}.
\item {\it Convergence}: Although the presence of a global Nash equilibrium has been proven, arriving at this equilibrium is not very simple. GANs frequently launch oscillating or cyclical behavior and are prone to converge to a local Nash equilibrium, which can be subjectively far from the global equilibrium.
\end{itemize}
The following sections present an overview of GAN architectures to address the above problems. In this paper, we review the synthetic image generation and identify different networks, wherein the earliest paper was proposed in 2014 \cite{goodfellow2014generative}. These techniques are grouped based on five main criterias: {\it network design}; {\it learning strategies}; {\it supervised methods}; {\it unsupervised methods}; {\it domain adaptation} and other methods (see Tabel~\ref{table:nonlin}).  As illustrated in Figure~\ref{fig:s1b} generation of synthetic images is one of active area of research in the recent years. In this review, we only focused on state-of-the-art image synthesizes techniques, and more than 200 research contributions are included. In addition, We used published codes by authors on {\fontfamily{cmr} \selectfont Github} to reproduce some results that are shown in this paper. We finish this survey by identifying directions for the future works.
%(\url{http://www.elsevier.com/locate/latex}).
%The class may be moved or copied to a place, usually,\linebreak
%\verb+$TEXMF/tex/latex/elsevier/+, %$%%%%%%%%%%%%%%%%%%%%%%%%%%%%
%or a folder which will be read                   
%by \LaTeX{} during document compilation.  The \TeX{} file
%database needs updation after moving/copying class file.  Usually,
%we use commands like \verb+mktexlsr+ or \verb+texhash+ depending
%upon the distribution and operating system.
%\section{Front matter}
%
%The author names and affiliations could be formatted in two ways:
%\begin{enumerate}[(1)]
%\item Group the authors per affiliation.
%\item Use footnotes to indicate the affiliations.
%\end{enumerate}
%See the front matter of this document for examples. 
%You are recommended to conform your choice to the journal you 
%are submitting to.
To design the first architecture of GAN, fully connected (FC) neural networks are used for both $G$ and $D$ \cite{goodfellow2014generative} to generate fake images based on Toronto Face Dataset \footnote{\fontfamily{qcr}\selectfont \color{purple}{https://inclass.kaggle.com/c/facial-expression-prediction2}}, MNIST \cite{lecun1998gradient} and CIFAR-10 \cite{krizhevsky2009learning}. Chen and Jiang \cite{chen2018building} proposed a GAN framework based on FC layers for building occupancy modeling which only shows high performance on few sets of data distributions. Later, other types of GAN are proposed that we discuss them in the following sections.

\subsection{Convolution GANs}
Moving from $FC$ to convolutional neural networks (CNNs) is suitable for the image data. Previous experiments have shown that it is extremely difficult to train $G$ and $D$ while using CNNs, mostly due to five reasons: Non-convergence, Diminished gradient, Unbalance between the generator and the discriminator, model collapse, hyper parameter selections. One solution is to use Laplacian pyramids of adversarial networks \cite{denton2015deep}. In this model, a real image is converted into a multi scale pyramid image, and a convolutional GAN is trained to produce multiscale and multi-level feature maps where the final feature map can be derived by combining all of them. The Laplacian pyramid is a linear invertible image demonstration containing band-pass images and a low-frequency residual.
\begin{equation}
\ h_k = \ell_k (I) = G_k (I) - u(G_{k+1} (I)) = I_k - u(I_{k+1})
\label{eq:e2}
\end{equation} 
where $k$ is the number of levels in the pyramid, $I$ denotes the image and $u (.)$ is the upsampling operator that smooths and expands $I$ to the double size. Hence $u (I)$ is a new image size. The factors $h_k$ at different $k$ level of the Laplacian pyramid $\ell_k (I)$ are created by computing the variance between adjacent levels in the Gaussian pyramid, $h_k = I_k$. To reconstruct an image from a Laplacian pyramid coefficients $[h_1, h_2, . . . , h_K]$ is achieved by applying the regressive recurrence and started with $I_k = h_k$ as follows:  
\begin{equation}
\ I_k = u (I_{k+1}) + h_k
\end{equation}
Following training, the model has a set of generative convent models $\{G_0, G_1, …,G_K\}$, while each of them captures the distribution of coefficients $h_k$ for images at a different scale of the network.
\begin{equation}
\overline I_k = u(\overline I_{k+1}) + \overline h_k = u(\overline I_{k+1}) + G_k(z_k, u(I_{k+1}))
\end{equation}
At the initial step the model starts by $\overline I_{k+1} = 0$ and at the final level $G_k$ produces a residual image $\overline I_k$  by expending noise vector $z_k:\overline I_k = G_k (z_k)$. Radford et al. \cite{radford2015unsupervised} introduced a deep convolutional GAN that enables smooth training for both G and D. This model uses the stride and fractionally-stride convolution layers which support the spatial down and up sampling operators to be significantly learned throughout the training. The role of these operators is to manage the changes in sample distributions and rates. For the 3D synthesize data generation, Wu et al. \cite{wu2016learning} presented an architecture that uses auto-encoder and long-range context information to directly reconstruct a 3D objects from a 2D input images. However, this work suffers from high computational cost. Guibas et al. \cite{guibas2017synthetic} proposed a new, two stage model by using dual network for generating synthetic medical images. Despite, the model has a lightweight network but the results are limited and the network is trained on a small size dataset.
\begin{table*} 
\caption{Tools available to researchers for addressing a synthesis image problem.}
\label{table:nonlin}
\resizebox{\textwidth}{!}{%
\begin{tabular}{p{.22\textwidth}|p{.3\textwidth}p{.2\textwidth}p{.22\textwidth}p{.2\textwidth}p{.2\textwidth}}
\hline
\multicolumn{6}{c}{Methods used in GANs for Image Synthesis Generation} \\ [1ex]
\hline
Network Design & Learning Strategies & Supervised Methods & Unsupervised Methods & Domain Adaptation & Other Methods\\  [1ex]
\hline
Residual Learning \cite{he2016deep} & Curriculum learning \cite{doan2019line} & cGAN \cite{isola2017image} & SD-GAN\cite{benaim2017one} & ADDA \cite{tzeng2017adversarial} & Context Fusion \cite{wu2016learning} \\ [1ex]
Recursive Learning \cite{wang2017shape} & Multi-supervision \cite{wang2019weakly} & PLDT \cite{yoo2016pixel} & DTN \cite{taigman2016unsupervised}& CycleGAN \cite{zhu2017unpaired} & Multitask learning \cite{mao2017aligngan} \\[1ex]
Multi-path Learning \cite{qi2020loss} & Loss functions: $\downarrow$ & SDF-MAN \cite{pu2019sdf}& DualGAN \cite{yi2017dualgan} & DiscoGAN \cite{kim2017learning}& Self- Ensemble \cite{mordido2018dropout} \\[1ex]
Channel Attention \cite{ma2018gan} & ~ ~ ~ ~ $\circ$ Content loss \cite{azadi2018multi} & CC-GAN \cite{denton2016semi} & In2i \cite{perera2018in2i} & AugGAN \cite{huang2018auggan} & Network interpolation \cite{achlioptas2018learning} \\ [1ex]
Advanced Convolution \cite{lutz2018alphagan} & ~ ~ ~ ~ $\circ$ Adversarial loss \cite{ulyanov2017takes} & FusedGAN \cite{bodla2018semi} &IR2VI \cite{liu2018ir2vi} & CoGAN \cite{liu2016coupled} & Feature Constancy \cite{qi2020loss} \\ [1ex]
Pixel  Learning \cite{wang2019weakly} & ~ ~ ~ ~ $\circ$ Cycle Consistency loss \cite{wang2018high} & GONet \cite{hirose2018gonet} & SCH-GAN \cite{zhang2018sch} & DISE \cite{chang2019all} & Distance Constraint \cite{radford2015unsupervised} \\ [1ex]
Pyramid Pooling \cite{shamsolmoali2020road} & ~ ~ ~ ~ $\circ$ TV loss \cite{lin2018pacgan} & GraphSGAN \cite{ding2018semi} & UNIT \cite{liu2017unsupervised} & CatGAN \cite{wang2017catgan}& $\cdots$ \\ [1ex]
$\cdots$ & ~ ~ ~ ~ $\circ$ Prior based loss \cite{xu2018attngan} & SCH-GAN \cite{zhang2018sch}	& XGAN \cite{royer2020xgan} & CGAN \cite{hong2018conditional} & $\cdots$ \\ [1ex]
$\cdots$ & ~ ~ ~ ~ $\circ$ Variation loss \cite{jang2018video} & SLSR \cite{ainam2019sparse} & CCD-GAN \cite{gomez2018unsupervised} & CDADA \cite{teng2019classifier} & $\cdots$ \\ [1ex]
$\cdots$ & ~ ~ ~ ~ $\circ$ Identity Preserving Loss \cite{huang2017face} & SS-GAN \cite{sricharan2017semi}& GANVO \cite{almalioglu2019ganvo} & Improved DTN \cite{polyak2018unsupervised} & $\cdots$ \\ [1ex]
$\cdots$ & ~ ~ ~ ~ $\circ$ Dual Learning Loss \cite{zhao2017dual} & $\cdots$ & GcGAN \cite{fu2019geometry}& CyCADA \cite{hoffman2018cycada} & $\cdots$ \\ [1ex]
$\cdots$ & ~ ~ ~ ~ $\circ$ Style Loss \cite{jing2019neural} & $\cdots$  & GM-GAN \cite{ben2018gaussian} & $\cdots$ & $\cdots$ \\ [1ex]
$\cdots$ & ~ ~ ~ ~ $\circ$ Pixel Loss \cite{pang2018visual} & $\cdots$  & WaterGAN \cite{li2017watergan} & $\cdots$ & $\cdots$ \\ [1ex]
$\cdots$ & ~ ~ ~ ~ $\circ$ Texture Loss \cite{liao2019unsupervised}  &  $\cdots$ & Vos-GAN \cite{spampinato2019adversarial}& $\cdots$ & $\cdots$ \\ [1ex]
\hline
\end{tabular}
}
\end{table*}
\subsection{Conditional GANs}
In \cite{isola2017image}, the authors proposed conditional GANs as a solution for image-to-image translation problems. The proposed model not only learns the mapping from input image to output image, but also adopted a loss function to train this mapping. This approach provides the opportunity to apply the same generic method to the problems that traditionally would need complex loss formulations. The architecture is shown in Figure~\ref{fig:s2}(b). As compared to the other GAN architectures, the conditional GANs have significant performance on the multi-modal data in comparison with \cite{zhu2017unpaired, zhu2017toward, zhao2016energy}. On the other hand, InfoGAN \cite{chen2016infogan} was another development that uses the mutual information between a small subset of the latent variables to gain semantic information. 
The architecture is presented in Figure~\ref{fig:s2}(c). Such model can be applied to determining different objects in an unsupervised way and also all the produced samples by InfoGAN are semantically well meaningful.
\begin{figure}[b]
\centering
  \includegraphics[width=3in]{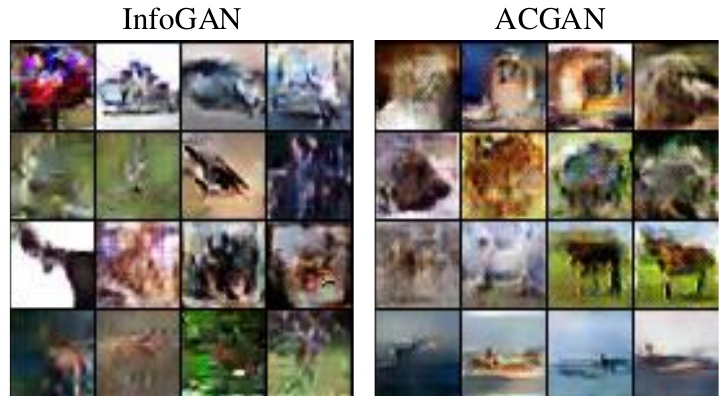}
  \caption{Sample generated results by InfoGAN and ACGAN on CIFAR-10}
\label{fig:info-AC}
\end{figure}

Zhou et al. \cite{zhou2018lp} introduced a normalization technique with conditional GAN that limits the searching space of the weights in a low-dimensional manifold. In \cite{lan2018demand, dai2017calibrating}, the authors proposed a conditional adversarial network for energy management systems. Their method is demonstrated to converge faster in term of number of epoch, but the authors did not highlight the model complexity. Odena et al. \cite{odena2017conditional} proposed a novel GAN classifier (\verb|ACGAN|)in which the architecture is similar to Infogan. In this model, the condition variable $c$ will not be added to the discriminator, and an external classifier is applied to predicting the probability over the class labels. The loss function is optimized to improve the class prediction.
In \cite{mariani2018bagan}, the authors proposed a data augmentation with balancing GAN (BAGAN) the architecture shows in Figure~\ref{fig:s2}(d). Class conditioning is applied in the hidden space to run the generation procedure towards the objected class. The $G$ in the BAGAN is adjusted with the encoder module that enables it to learn in the hidden space. The structure of BAGAN is similar to InfoGAN and ACGAN. However, BAGAN only generate a single output but, InfoGAN and ACGAN have two outputs.\par
 In \cite{karacan2016learning}, the author presented a deep conditional GAN model that takes its strength from the semantic layout and scene attributes integrated as conditioning variables. This approach able to produces realistic images under different situations, with clear object edges. Figure \ref{fig:info-AC} compares the generated images by InfoGAN and ACGAN on CIFAR-10.

%%%%
\subsection{Auto-Encoders GAN}
The auto-encoders networks have great performance in different computer vision tasks. Such networks are generally composed of one encoder and one decoder.
In such networks, the model learns a deterministic mapping via the encoder and the decoder. Combination of these two modules helps the network to reconstruct images that look similar to the original one. 
 \begin{figure*}
 \centering
  \includegraphics[width=7in]{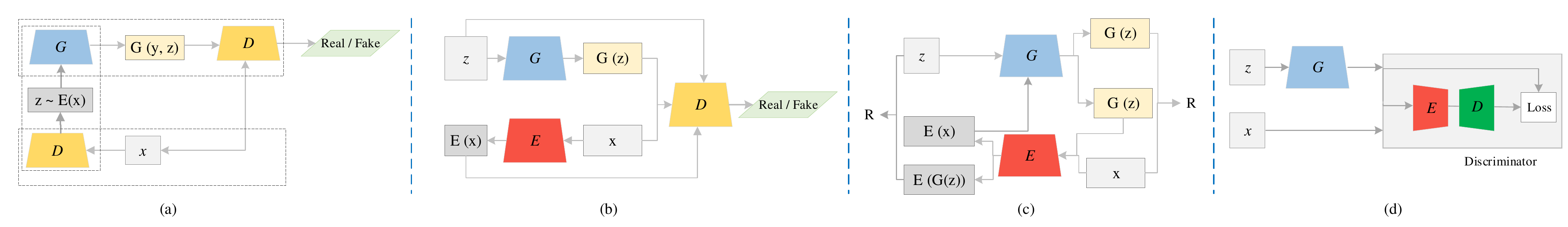}
  \caption{Architecture of {\bf (a)} auto-encoders GAN; {\bf (b)} BiGAN {\bf (c)} Adversarial Generator-Encoder Network (AGE) \cite{ulyanov2017takes}; {\bf (d)} BEGAN architecture \cite{berthelot2017began}. $z$ is the random noise for $G$ and $x$ is input image. BEGAN deploys an auto-encoder architecture for the discriminator.}
\label{fig:s4}
\end{figure*}
%%%%%%
Auto-encoders generally for learning follow the non-linear mappings in both directions. In \cite{larsen2016autoencoding}, the authors suggested to combine auto-encoder \cite{kingma2013auto} with GAN \cite{goodfellow2014generative} to collect the advantages of both the models, where GAN can produce sharp images however lost some features. On the other hand, images generated by auto-encoders \cite{kingma2013auto} are blurry but the model is efficient and accurate. Figure~\ref{fig:s4}(a), illustrates the architecture of the adversarial auto-encoders. In this model, the auto-encoder part normalizes the encoder $E$ by regulating the distribution $(z\sim N(0,1))$, and the model loss formulated as Eq.(\ref{eq:e5}): 
\begin{equation}
\ell_{AEGAN} = - E_{z \sim q(x|x)} \log [p(x|z)] + D_{KL} (q(z|x)||p(x))
\label{eq:e5}
\end{equation}
%%%%%%%%
%\begin{figure}[b]
%\centering
%  \includegraphics[width=2.5in]{fig5}
%  \caption{Architecture of : {\bf (a)} Adversarial Generator-Encoder Network (AGE) \cite{ulyanov2017takes}; {\bf (b)} BEGAN architecture \cite{berthelot2017began}. $z$ is the random noise for $G$ and $x$ is input image. BEGAN deploys an auto-encoder architecture for the discriminator.}
%\label{fig:s5}
%\end{figure}
while $z \sim E(x) = q(z|x)$ ; $x \sim G(z) = p(x|z)$ and $D_{KL}$ is the Kullback-Leibler divergence. Moreover, auto-encoder GAN \cite{larsen2016autoencoding} used the reconstruction loss of Auto-Encoder for $D$ to check how similar the generated samples looks against the real samples and also guide the generator to produce more similar samples.
Donahue et al. \cite{donahue2016adversarial} introduced a Generative Adversarial Networks called (BiGAN), to produce valid inferences. Figure~\ref{fig:s4}(b) shows the architecture of BiGAN. An encoder is added to these models, in addition to the generator and discriminator, to map the generated data back to the latent space. This approach uses the encoder in the discriminator as a feature capture tool.  In \cite{dumoulin2016adversarially}, the authors proposed an Adversarial Learned Inference (ALI) that uses the encoder for latent learning. The proposed model is efficient and a wide range of evaluations is presented. These autho-encoder structures suffers from the generation results. \cite{donahue2016adversarial} and \cite{dumoulin2016adversarially} are proposed to address the problem of uninformative latent space in the GAN technique by matching the joint distributions of the real data and latent variable, however, the objective functions do not rely on the relationship between the latent variables and the observations, thereby they couldn't achieve a faithful reconstruction performance.
\\

Ulyanov et al. \cite{ulyanov2017takes} proposed a model that uses the encoder in the generator "{\itshape Adversarial Generator-Encoder Network} (AGE)", in which the adversarial loss is applied between the generator and the encoder, and the network does not require the discriminator. Figure~\ref{fig:s4}(c) demonstrates the AGE architecture in which R denotes the reconstruction loss function. In this model, the aim of the generator is to reduce the gap between the latent distribution {\itshape z} and the synthetic data distribution, while the aim of the encoder is to maximize the divergence between {\itshape z} and {\itshape E(G(z))}. Additionally, the models use the reconstruction loss function to avoid the possibility of mode collapse. One interesting point in the \cite{ulyanov2017takes} is that, if the encoder-decoder models are not explicitly trained for reconstruction task, this can be done by projecting data samples into the latent space through the encoder, then pass the intermediary points via decoder and projecting them back to the original data space. In this case, the reconstructions often retain some semantic features from the original data, while perceptually are different from the original ones. 
In \cite{luo2017learning}, the authors introduced a new approach based on inverse $G$ as the encoder and pre-trained $G$ as the decoder of an auto-encoder network to train the inverse $G$ model. In this model, the gap between the input and the output images that are produced by pre-trained auto-encoder’s GANs is directly minimized. 
\\

Moreover, Lutz et al. \cite{lutz2018alphagan} introduced the first GAN for natural image matting. This network is trained to predict appealing alphas by applying dilated convolution into encoder-decoder architecture to capture global contextual information and adopting the adversarial loss to improve the performance of the model in classification of the composited images. 
\begin{equation}
\ p(D(x)|z) =  N \Big(D(x)|D(\hat {x}), I \Big)
\end{equation}
Boundary Equilibrium GAN "BEGAN" as shown in Figure~\ref{fig:s4}(d) adopted the auto-encoder model for the discriminator \cite{berthelot2017began}.
In comparison with the standard optimization, BEGAN produces the auto-encoder loss distributions by using a loss extracted from the Wasserstein distance. This optimization supports G to produce more realistic data. Since the synthetic data distribution is very close to {\itshape 0} and the real data distribution has not been used to lean the model, therefore D at early stage is not able to well distinguish the generated samples. Moreover, the reported set of images for experimentation is small.
\begin{figure}[t]
  \centering
  \includegraphics[width=3in]{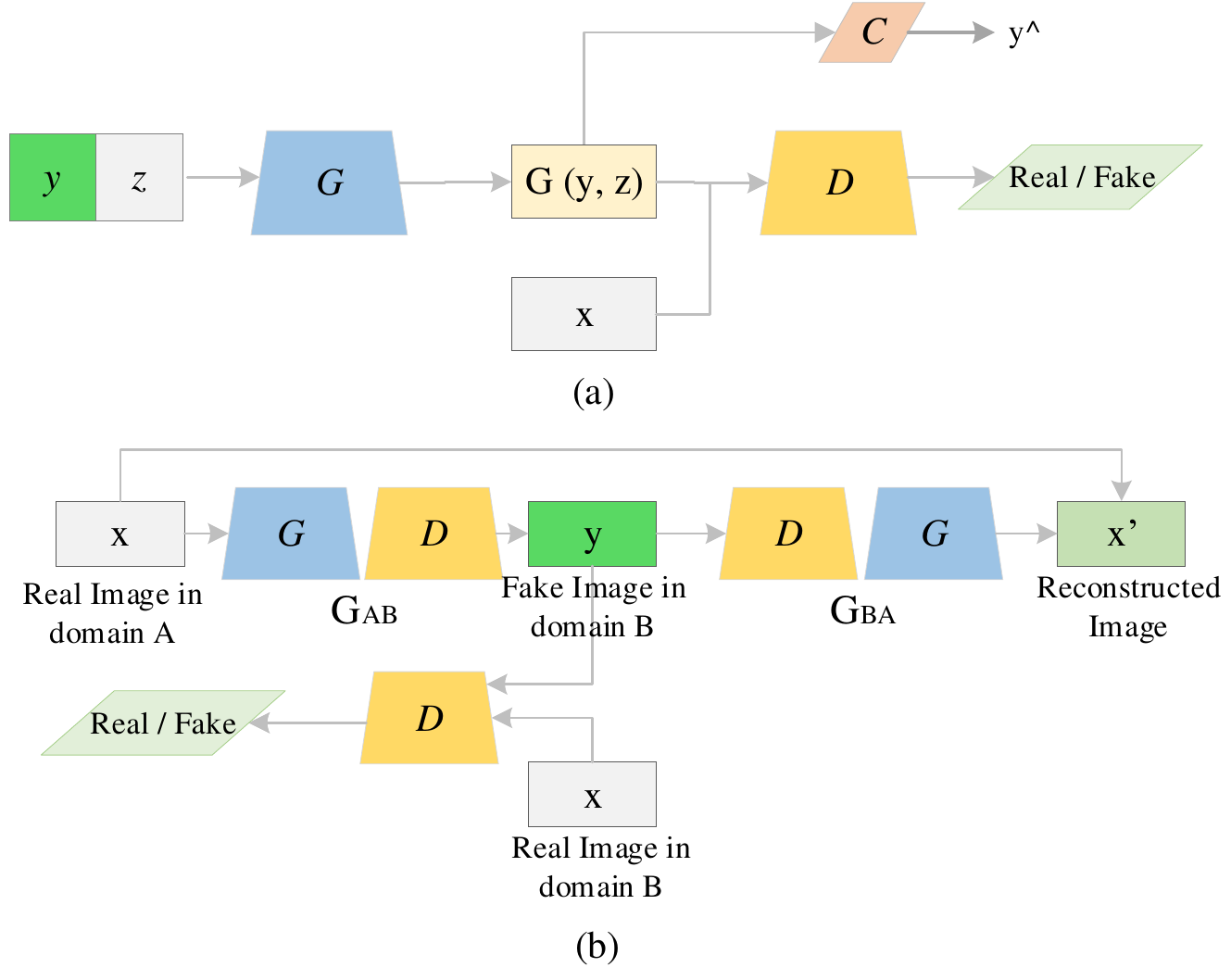}
  \caption{Architecture of: {\bf (a)}  GAN with auxiliary classifier \cite{odena2017conditional}, in which 'y' is the conditional input label and C is the classifier; {\bf (b)} CycleGAN \cite{zhu2017unpaired}.}
\label{fig:s6}
\end{figure}

\subsection{Progressive and Auxiliary Classifier GAN }
In progressive GANs, the model expands the architecture of the standard network \cite{karras2017progressive} where the idea was extracted from progressive neural networks \cite{rusu2016progressive}. This model has high performance as it can receive additional leverage via lateral connections to earlier learned features. This architecture is widely used for extracting complex features. For training, the model starts with low resolution images and progressively $G$ and $D$ grow to reach the desirable results. It is worth of mentioning, during this growing process, all the variables remain trainable. This progressive training strategy helps the networks to be of stable learning. Currently, several state-of-the-art GANs adopted such training strategy to improve their overall performance \cite{shamsolmoali2019g, zareapoor2019perceptual}. In \cite{heljakka2018pioneer}, the authors adopted the progressive GANs into Autoencoder network for image reconstruction. The authors claim this model has promising results in image synthesis and inference. However, the model is only evaluated in CelebA dataset and the efficiency of the proposed model is not evaluated.

%\subsection{Self-attention GAN}
%The ordinary style CNNs may only extract local spatial features and the receptive field cannot cover sufficient structures, which is the weakness of CNN-based GANs that faced learning difficulty from multi-class image datasets and the main points in image producing may shift, for example, the nose may not place in the correct position. A self-attention approach is proposed that helps the network to have large receptive field without affecting the computational efficiency \cite{ma2018gan}. Self-attention GAN adopts this structure in both the generator and the discriminator that enable the networks to learn from long-range and global features. 

%\subsection{GAN with Auxiliary Classifier}
In order to boost up the performance of GAN for semi-supervised learning, \cite{odena2017conditional, dash2017tac} proposed to add an additional precise auxiliary classifier to the discriminator. Figure~\ref{fig:s6}(a) represents the architecture of the auxiliary GAN, in where C denotes the auxiliary classifier. Auxiliary classifiers provide the pre-trained modules (the network can be pre-trained on the big datasets such as ImageNet). The results show an auxiliary classifier GAN can generate sharper edge images with the ability in handling the collapse problem. The GANs with auxiliary classifiers had significant performance in applications such as image-to-image translation \cite{odena2017conditional} and text-to-image synthesis \cite{dash2017tac}. More examples is given in Table~\ref{table:nonlin}. 
\begin{figure}[b]
  \centering
  \includegraphics[width=2.8in]{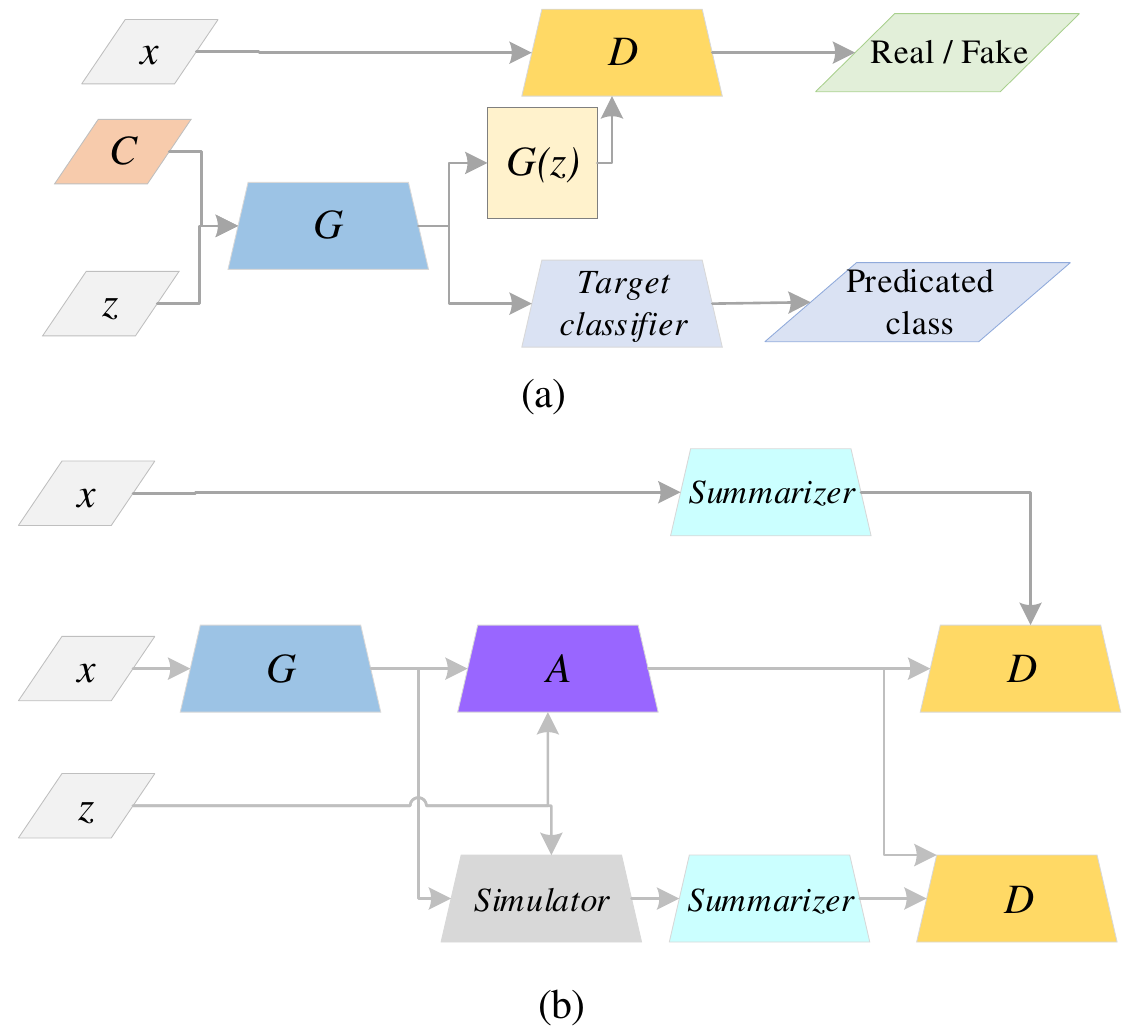}
  \caption{Architecture of;  {\bf (a)} class-condition GAN \cite{tsai2018customizing}; {\bf (b)} optimized GAN \cite{qi2020loss}. The model contains two different optimization networks.}
\label{fig:s7}
\end{figure}
\subsection{Adversarial Domain Adaptation}
Unpaired image-to-image translation models recently have superior performance on different domain adaptation tasks. These type of methods such as, ADDA \cite{tzeng2017adversarial}, CycleGAN \cite{zhu2017unpaired}, CyCADA \cite{hoffman2018cycada}, DiscoGAN \cite{kim2017learning}, AugGAN \cite{huang2018auggan}, and DualGAN \cite{yi2017dualgan} have almost the same architecture that makes unpaired image-to-image translation by introducing the sequence consistency. In addition, Tsai \cite{tsai2018customizing} to increase the visual quality of generated images proposed class-conditional GANs. This model uses a non-perturbation based framework that produces adversarial examples from class-conditional GANs. Therefore, the generated data will not resemble any similar data and consequently enlarge example diversity and difficulty in adversarial defense. The architecture of class-condition GANs shown in Figure~\ref{fig:s7}(a). 

\begin{figure*}[t]
  \centering
  \includegraphics[width=6.8in]{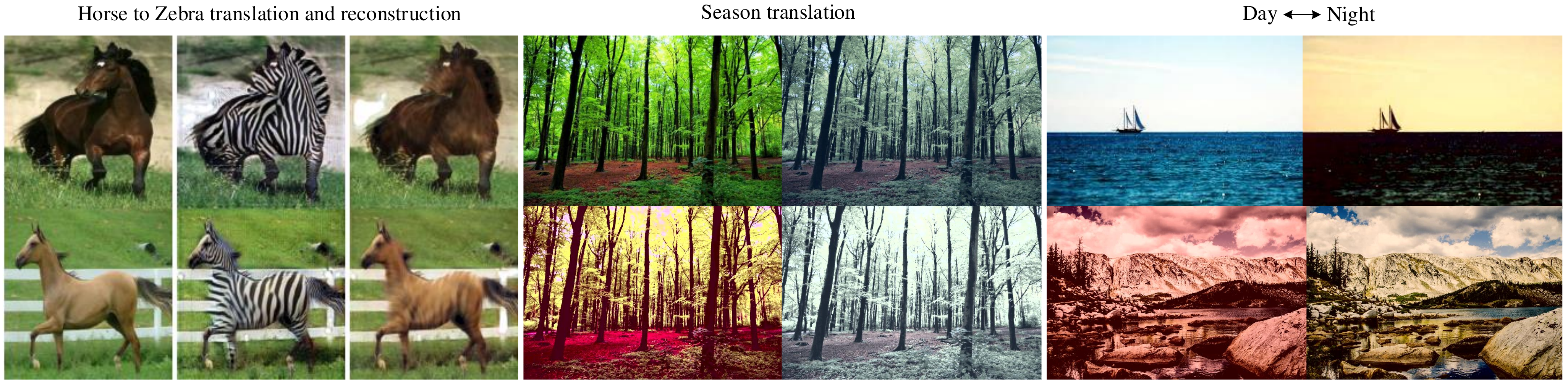}
  \caption{Sample generated results using CycleGAN \cite{zhu2017unpaired} \protect \footnotemark}
\label{fig:s8}
\end{figure*}

\begin{figure}[hb]
  \includegraphics[width=3.5in]{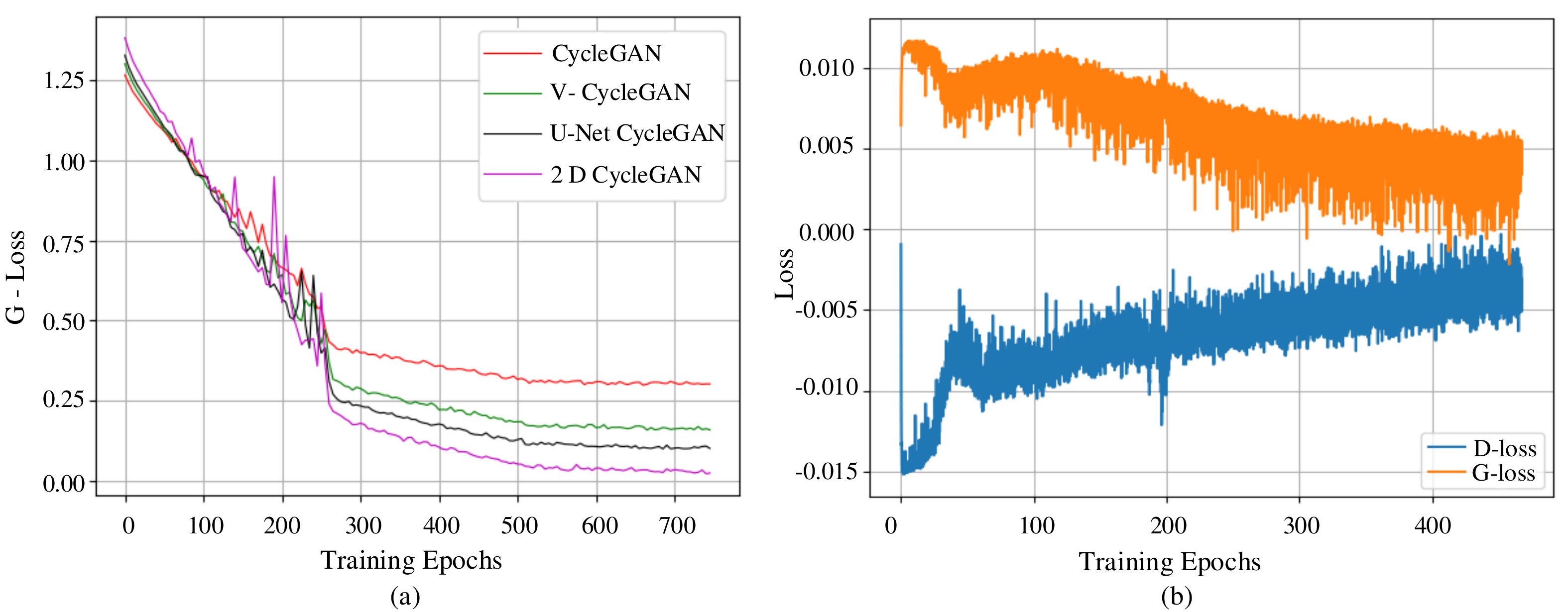}
  \caption{Performance of CycleGAN \cite{zhu2017unpaired}and DCGAN \cite{radford2015unsupervised} based on their training loss. The models are evaluated on flickr dataset \protect \footnotemark.  {\bf (a)} Training loss of CycleGAN on different setting;  {\bf (b)} DCGAN training loss of G and D. }
\label{fig:s9}
\end{figure}

Recently, there is a new model introduced for unpaired images called CoGAN \cite{liu2016coupled} that the authors proposed to use two shared-weight generators for producing images of two domains with random noise. All these models have convincing visual results on numerous image-to-image translation tasks, nonetheless, large domain shift may degrade the ability of these methods for generating large-scale training data. The authors claim efficient training speed along with better image quality. Figure~\ref{fig:s6}(b) represents the architecture of cycleGAN. Chang et al. \cite{chang2019all} introduced a domain adaptation adversarial network for unsupervised segmentation, which tries to handover the information learned from synthetic datasets with ground-truth labels to real images without any annotation \cite{liao2019unsupervised}.
The authors proposed a domain adaptation method to separate images into domain-invariant texture and domain-specific structure, which can further be used for image translation across domains and helps label transfer to improve segmentation accuracy. Figure~\ref{fig:s8} shows some sample results generated by CycleGAN \cite{zhu2017unpaired}. There have been a wide range of efforts for improving and applying CycleGAN in other domains. In \cite{zhu2017toward}, the authors proposed to change the structure of the network and adopting U-Net \cite{shamsolmoali2019novel} to generate more realistic images, and in \cite{lu2018attribute} dual discriminator is applied in the network to effectively diversify the estimated density in capturing multi-modes. Figure~\ref{fig:s9}(a) represents the training loss of CycleGAN on different setting.  
\footnotetext{{\fontfamily{qcr}\selectfont \color{purple}{https://github.com/junyanz/CycleGAN}}}
Wang and Zhang \cite{wang2017catgan} introduced an efficient solution for transfer GAN to domain alignment. The basic principles of proposed model focus on the domain generation and adaptation. The proposed network comprises of two slim and symmetric sub-networks, which then formulates a coupled adversarial learning framework. Hong et al. \cite{hong2018conditional} introduced a novel structured domain adaption model for multiple image semantic segmentation, this work integrates GAN into the FCN to minimize the source and target domain gaps. Specially, the generator is learned to convert features of synthetic images to be quite similar to a real-image features, and a discriminator is trained to differentiate the real images from the generated ones. 
\\

Teng et al. \cite{teng2019classifier} presented a classifier-constrained adversarial domain adaptation model for cross-domain semi-supervised classification which only tested on a small size remote sensing image dataset. Polyak et al. \cite{polyak2018unsupervised} proposed an unsupervised domain adaptation GAN which is based on discrepancy to handle the problems regarding image mapping between different domains. In \cite{jethava2017easy}, the authors proposed a GANs model that has high accuracy performance on high-dimensional data. An approximator network is adopted into the GANs architecture to generate a rich set of features. Therefore, the model has high scalability and ability to handle complex probability distributions. The architecture this model shown in Figure~\ref{fig:s7}(b). Wang et al. \cite{wang2019weakly} proposed a weakly supervised adversarial domain adaptation to improve the segmentation performance from the synthetic data, which consists of three networks. The detection model focuses on detecting objects and predicting a segmentation map, the pixel-level domain classifier attempts to distinguish the domains of image features, an object-level domain classifier discriminates the objects and predicts the objects classes. On the other hand, in \cite{mao2017aligngan} the author proposed a model to enhance the ability of the conditional GANs. The proposed model evaluated on large set of datasets and multiple tasks. The experimental results show that this model has the ability to successfully align the cross-domain images without paired samples. Moreover, \cite{mokhayeri2020cross} proposed a cross-domain regulated GAN for synthetic 3D face generation.
Table ~\ref{table:nonlin}, describes the set of tools available to a researcher for addressing a synthesis image problem.

\footnotetext{{\fontfamily{qcr}\selectfont \color{purple}{https://www.flickr.com/photos/aaefros}}}
\section{GANs and Loss-Variants}
\label{sec:3}
\begin{figure}[t]
\hspace{-1cm}
  \includegraphics[width=3.45in]{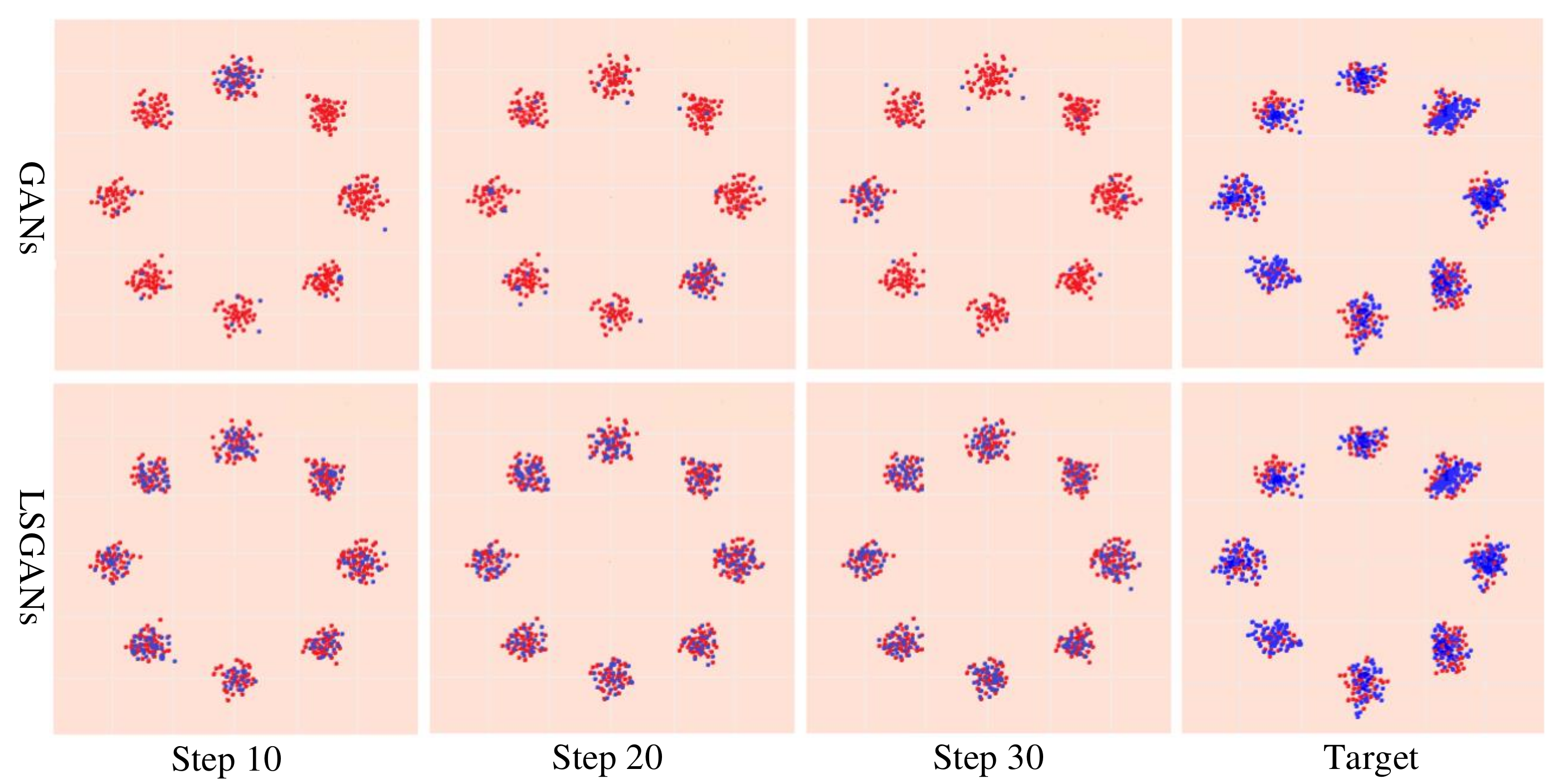}
  \caption{Dynamic results of Gaussian kernel for regular GANs and LSGANs on {\fontfamily{cmr} \selectfont CIFAR-10} dataset. \textcolor{red}{Red} portions represent real samples and \textcolor{blue}{Blue} portions signify generated ones. GAN almost in all the steps generated the synthetic samples that are very close to their corresponding original (target) samples.}
\label{fig:s11}
\end{figure}

Changes in the loss functions resulted in GANs Loss-variant. Several works \cite{qi2020loss, wan2020deep} have already improved the optimality and the convergence of GANs training, but still unstable training is the major issue of GAN models.
 The fault is initiated by the global optimality which is pointed in \cite{ma2020pan, zhan2019spatial}. The global optimality is determined while an optimal $D$ is gained for a particular $G$. The optimal $D$ can be achieved, whenever the derivative $d$ of the loss in Eq.(\ref{eq:e2}) equals to $0$.  Consequently we have,
\begin{equation}
-\frac{p_r (x)}{d(x)} + \frac{p_g (x)}{1- d(x)} = 0 \  ; 
\  D^* (x) = \frac{p_r (x)}{p_r (x) + p_g (x)} 
\end{equation}
in which {\itshape X} is real data, $D^* (x)$ is the optimized discriminator, $p_r (X)$ is the distribution of real data and $p_g (X)$ is the distribution of the generated data over real data {\itshape X}. By having the optimal D, the loss of G can be rephrased by adopting $D^* (X)$ into Eq.(\ref{eq:e2}) : 
\begin{equation}
\begin{aligned}
\ell_G &= E_{x\sim p_r}\log \frac{p_r (x)}{\frac{1}{2}[p_r (X) + p_g (x)]} + && \\
&E_{x\sim p_g}\log \frac{p_g (X)}{\frac{1}{2}[p_r (X) + p_g (x)]} - 2\log 2 &&
\label{eq:e8}
\end{aligned}
\end{equation} 
Eq.(\ref{eq:e8}) shows the loss function of a GAN while having optimized the discriminator and it is based on two fundamental probability measurement metrics. 
One is Kullback-Leibler {\itshape KL}divergence which is formulated as: 
\begin{equation}
K\ell (p_1\Vert p_2)= E_{x\sim p_1} \log \frac{p_1}{p_2},
\end{equation}
Another is Jensen-Shannon {\itshape JS} divergence which is defined as:
\begin{equation}
JS (p_1\Vert p_2)= \frac{1}{2} K\ell (p_1\Vert \frac{p_1 + p_2}{2}) + \frac{1}{2} K\ell (p_2 \Vert \frac{p_1 + p_2}{2}). 
\label{eq:e10}
\end{equation}
thereby, the loss of $G$ based on the optimal $D^* (.)$ in (\ref{eq:e8}) can be stated as: 
\begin{equation}
\ell_G = 2JS (p_r\Vert p_g) - 2\log 2 
\label{eq:e11}
\end{equation} 
which shows that the average loss for G now resulted from the minimization of the JS divergence between $p_r$ and $p_g$. In \cite{wang2019generative} the authors extensively reviewed the GAN's loss functions.
\begin{figure*}
  \centering
  \includegraphics[width=6.5in]{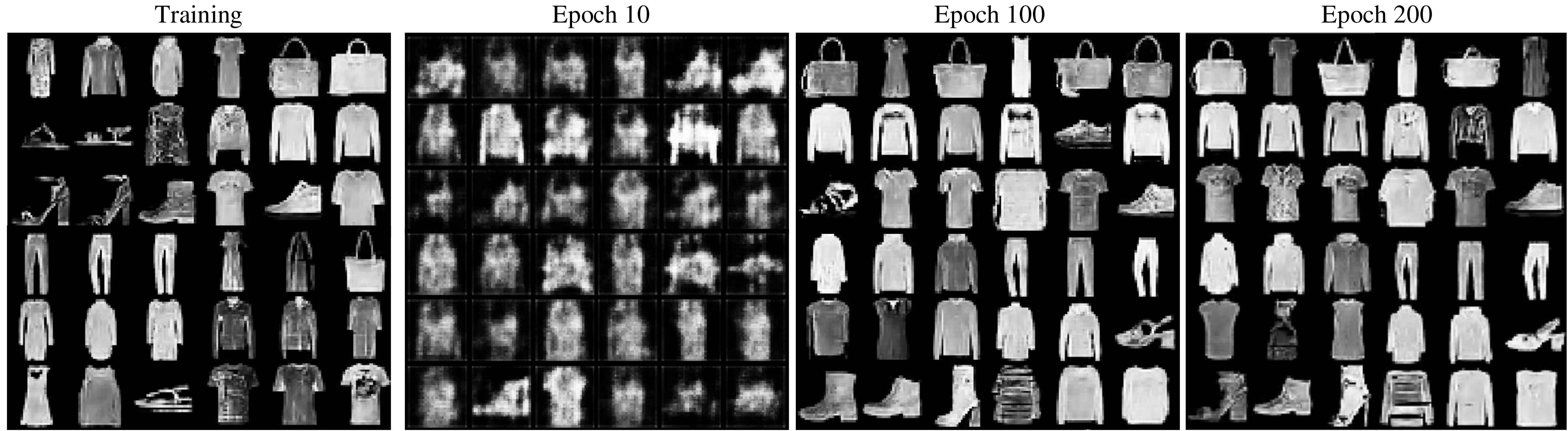}
  \caption{Sample training results of different iterations on FashionMNIST dataset using DCGAN \cite{radford2015unsupervised} \protect \footnotemark.}
\label{fig:s8b}
\end{figure*} 
\footnotetext{{\fontfamily{qcr}\selectfont \color{purple}{https://github.com/carpedm20/DCGAN-tensorflow}}}
In \cite{mao2017least}, the authors demonstrated that the optimization of least square GAN equals to decreasing the difference in the Pearson $\chi^2$ divergence between $p_r + p_g$ and $2p_g$ when {\itshape a}, {\itshape b} and $\alpha$ fulfill the condition of $b - c = 1$ and $b - a = 2$. Figure~\ref{fig:s11} illustrates the dynamic results of Gaussian kernel density estimation. It can be observed; regular GANs suffer from mode collapse. They generate samples nearby a single valid mode of the data distribution. On the other hand {\itshape LSGAN} \cite{mao2017least} learns the Gaussian mixture distribution successfully. Moreover, Zhou et al. \cite{zhou2017activation} statistically study the details of the existing variants of GANs that take the advantages of class label information. Based on class aware gradient and cross-entropy decomposition, the authors proposed a model called Activation Maximization GAN as a solution to enhance the GAN training.

A wide range of experiments have been conducted to evaluate the efficiency effectiveness of the model.
Qi \cite{qi2020loss} introduced loss sensitive GAN to train a generator to create more realistic images by minimizing the boundaries between the real and fake samples. The author states that, the training issues such as model collapse and vanishing gradient influencing the original GAN with a non-parametric hypothesis that the discriminator has the ability to detect any type of distribution between the real and fake samples. As previously discussed, sometimes there is an overlapping between the real and fake samples distribution which is ignorable. In the loss sensitive GAN, the classification task of $D$ is limited and is learned by a loss function $L_\theta$ parameterized with $\theta$, which expects the real samples to be of smaller loss than fake ones. For the training of this loss function, the following limitations are applied:
\begin{equation}
\ell_\theta (x) \leq \ell_\theta (G(z)) - \Delta (X, G(z))
\end{equation} 
in which $\Delta(X, G(z))$ determines the divergence gap between the real and fake samples. This restriction results in the separation of a real sample from the generated ones by a margin of $\Delta(X, G(z))$. The loss sensitive GAN is optimized as follows:
\begin{equation}
\begin{aligned}
\min_{D}(\ell_D) = E_{x \sim p_r (x)}\ell_\theta (X) +  \\
\lambda E_{x \sim p_r}\Big(\Delta(X, G(z)) + \ell_\theta (X) - \ell_\theta(G(z))\Big), \\
\min_{G}(\ell_G) = E_{z \sim p_z}\ell_\theta (G(z)). 
\label{eq:20}
\end{aligned} 
\end{equation}
while the balancing parameter $\lambda$ is positive then $(a) = max(a, 0)$. To the second term of Eq.(\ref{eq:20}), $\Delta (X, G(z))$ is added to optimize $D$ and prevent it from separating the real samples from the generated ones. Tolstikhin et al. \cite{tolstikhin2017adagan} proposed a method, called AdaGAN, in which every steps take a new component into a model by running a GAN process on a re-weighted sample, which motivated by boosting algorithms. The authors prove that progressive procedure leads to better convergence and the true distribution. The experiments illustrate that this procedure addresses the problem of missing modes. In \cite{liu2018latent, basioti2020designing}, the authors proposed a model which has GANs structure to convert a simple distribution to a data distribution in the latent space by training the discriminator between a simple distribution and a latent-space data distribution. Zhang et al. \cite{zhang2018generating} proposed an adversarial information maximization model, to handle the diversity issue of generated samples. The model regularly compares synthetic and real samples and accordingly updates the $G$ and $D$ to enhance the performance. 

\section{Datasets}
\label{sec:4}
The first step in building a GAN model is to collect enough numbers of training datasets for the discriminator to train the network. Once a sufficient amount of data has been collected, we can implement GAN for already existing datasets to learn and then apply the same implementation on the newly created dataset. In this section, we summarize some of the existing datasets that are widely used in evaluation of the current approaches. 
\begin{itemize}
\item {\verb|MNIST|}: this database is a handwritten digits that contains 60,000 training samples, and 10,000 testing samples. The size of each image is 28×28 and linearized as a vector of size 1×784. Therefore, the training dataset and test one have 2-d vectors of size 60000×784 and 10000×784 respectively.
\item {\verb|Fashion MNIST|}: this new dataset contains 60,000 samples for training set and 10,000 samples for testing. Each image sample has 28×28 grayscale size and in total there are 10 classes. 
\item {\verb|CIFAR-10|}: this dataset contains 50,000 training samples and 10,000 testing samples. Each color sample has 32×32 size and in total there are 10 classes, with 6,000 image samples per class. 
\item {\verb|CIFAR-100|}: this dataset is similar to \verb|CIFAR-10|, the difference is that, it has 100 classes and each containing 600 images. Each class contains 500 training and 100 testing samples. The images are well labeling and have high resolution.
\item {\verb|Celeb|A}: this dataset contains more than 200,000 celebrity face images, each sample has 40 attribute annotations and the images cover large pose variations and background clutter.  
\item {\verb|Cityscapes|}: this dataset consist of a various set of video sequences recorded in street from several cities, with high quality pixel-level annotations of 5,000 frames plus a larger set of 20,000 weakly annotated frames.
\item {\verb|Toronto Faces Dataset|}: this large dataset consist of $32\times32$ grayscale images that contain faces that have been take out from different sources. All subset of the faces have been well labeled. 
\item {\verb|UT Zappos50K|}: this dataset contains 50,025 catalogs of shoe images that collected from Zappos.com. These images are separated to 4 major categories “shoes, sandals, slippers, and boots”, functional types and the individual brands. 
\item {\verb|ImageNet|}: this image database is mainly collected to be used for  object recognition and contains more than 14 million images in 20,000 categories, which hand-annotated. 
\item {\verb|Van Gogh|}: this dataset consists of 1,000 Van Gogh paintings images that captured from the Amesterdam museum.  
\item {\verb|DSLR|}: this dataset consist of 22,000 images, containing 4549 photos from Sony mobile phone, 5727 from iPhone and 6015 photos from BlackBerry, for each mobile phone photo, there is a corresponding photo from the Canon DSLR. The photos have been captured in the daytime in a several places and in various illumination and weather conditions.  
\item {\verb|Caltech-UCSD Birds-200-2011|}: the dataset consist of 200 categories and in total 11,788 bird images. Each image has 15 Part Locations, 312 Binary Attributes, 1 bounding box.
\item{\verb|SVHN|}: contains more than 600,000 high resolution digit images which generally do not require any preprocessing and obtained from house numbers in Google Street. This dataset contains 73,257 training samples, 26,032 testing samples, plus 531,131 samples that can be used for extra training.
\item{\verb|FaceScrub|}: the dataset consists of 106,863 face images of 530 male and female celebrities, and in average 200 images from each person. 
\item{\verb|Paris StreetView|}: this dataset contains 6412 sample images that collected from Flickr by searching for the specific Paris monuments.
\item{\verb|YouTubeFace|}: this dataset contains  near to 3425 videos from 1595 different persons. These videos are downloaded from YouTube. The shortest clip length is 48 frames, the longest one is 6070 frames, and the normal length of the video clips are 181.3 frames.
\item{\verb|CartoonSet|}: this dataset contains cartoon face that composed of 16 modules including 12 facial features and 4 color features which are selected from a discrete set of RGB values. The dataset has 9000 cartoons.
\item{\verb|SENSIAC|}: this is a large-scale image dataset that used for object recognition. It consists of 207GB of middle-wave infrared videos and 106GB of VI videos that manually labeled. 
\end{itemize}
%%%%%
\section{Synthetic Image Generation Methods}
\label{sec:5}
In the last few years synthetic data is widely used to overcome the burden of generating large datasets for training CNNs. A wide variety of data synthesis methods have been proposed in literature, starting from photo-realistic image reproduction \cite{krahenbuhl2018free, richter2017playing, varol2017learning} and learning-based models for synthetic image generation \cite{salimans2016improved} to approaches for data augmentation that automated the operation for producing new sample images from actual training set \cite{gupta2016synthetic, ratner2017learning}. Conventional data augmentation methods have used image transformations that keep class labels \cite{tran2017bayesian}, while recent approaches \cite{ratner2017learning} introduced a more general image transformations method, involving image composition.
Synthetic data generation methods must generate data that have three significant features. It should be a) effective: produce meaningful and sufficient data samples, b) task-aware: create samples that contribute for better performance of target network, and c) realistic: produce realistic samples that assist in minimizing domain gaps and enhance generalization. In this section we briefly discuss the main methods that are used for synthetic image generation. 
%%
%%
%\begin{figure}[h]
%  \includegraphics[width=4in]{fig12}
%  \caption[Caption for LOF]{3D face generated samples by DCGANs \protect \footnotemark}
%\label{fig:s12}
%\end{figure}
%\footnotetext{{\fontfamily{qcr}\selectfont \color{purple}{https://github.com/carpedm20/DCGAN-tensorflow}}}

\subsection{Single-stage methods:}
The approaches in this section follow the standard form of GAN for just using a single G and D in their architectures, and G and D have simple networks without additional connections. Radford et al. \cite{radford2015unsupervised} introduced deep convolutional generative adversarial networks (DCGANs), which have some architectural constraints that have strong performance for unsupervised learning that has high efficiency and accuracy. Some sample training results of DCGAN on Fashion MNIST dataset are shown in Figure~\ref{fig:s8b}. Salimans et al. \cite{salimans2016improved}, proposed a mathematical model to improve the semi-supervised training of GANs and highly realistic. The model tested on wide range of datasets including ImageNet and was able to generate recognizable features. Zhao et al. \cite{zhao2017dual} proposed a cross-domain image captioning approach that uses a novel dual learning mechanism to overcome generating image problems of GANs. Lee and Seok \cite{lee2019controllable} introduced ControlGAN to manage the random distribution of produced samples by separating discriminator from classifier. In ControlGAN the generator is intended to produce synthetic images with the precise detailed. On the other hand, Mukherjee et al. \cite{mukherjee2019clustergan}, introduced ClusterGAN as a new mechanism for clustering by sampling latent variables from a mixture of one-hot encoded variables and continuous latent variables which has similar structure to ControlGAN. 
\begin{figure}[t]
\centering
  \includegraphics[width=3.4in]{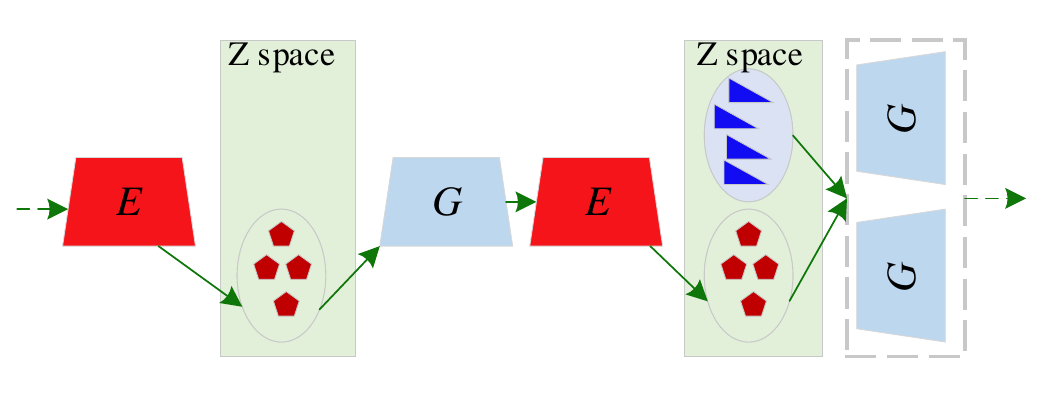}
  \caption{Performance of single-path and dual-path network.  Left: standard single path model to learn the latent representation. Right: Two-pathway network combined with self-supervised learning, which can learn complete representations.}
\label{fig:s13}
\end{figure}
\\

Zhong et al. \cite{zhong2020generative} also proposed a new model, called decoder-encoder generative adversarial networks, which take both the advantage of adversarial training and the variational Bayesain inference to increase the quality of generated images. Such methods are relatively convenient to implement as compared to multi-stage and hierarchy methods, and these methods generally achieve good results. Bhattarai et al. \cite{bhattarai2020sampling} proposed an efficient light weighted model for synthetic image generation utilizing the learning parameters during training. These include discriminator’s realism confidence scores and the confidence on the target label of the synthetic data. 
In addition, they adopted a reinforcement learning algorithm to automatically search a subset of meaningful synthetic examples from a large pool of generated data. In Figure~\ref{fig:s9}(b), we compare the training loss of G and D in DCGAN \cite{radford2015unsupervised}.

\subsection{Multi-stage methods:}
Opposite to the {\itshape single-stage} methods, the multi-stage methods use multiple generators and discriminators in their architectures, in which different generators are in charge of different tasks. The idea behind those approaches is to distinct an image into different portions, for example, {\bfseries "foreground \& background"} and {\bfseries "styles  \& structure"}. It is worth to mention that, the generators either works in a parallel or sequential way. Hong et al. \cite{hoang2018mgan} proposed to train the GANs with a combination of generators to prevail the mode collapse problem. The key idea is to use multiple generators, in place of a single one. This architecture, proven to be highly operational at covering diverse data modes, which handle the collapse problem.
Wang and Gupta \cite{wang2016generative} introduced to use two GANs. However, the authors show limited experiments and did not discuss the computation cost of the proposed model. The Structure-GAN \cite{deng2017structured} adopted the DCGAN \cite{radford2015unsupervised} structure with some modifications. In Structure-GAN \cite{deng2017structured}, the created samples and the noise vector feed to some convolution layers, and at the end the results are merged to create one single tensor which will goes through other transport layers. Structure-GAN \cite{deng2017structured} composed of fully-connected network that changes an image into normal map. The only drawback of the \cite{deng2017structured} is that it requires using additional features to gain ground-truth for the surface mapping.  Tian et al. \cite{tian2018cr} introduced CR-GAN which, besides the single reconstruction path, has a generator unit to maintain the learning. These two learning pathways are equally collaborate for sharing the parameters for improving the generalization ability. 
\\

\begin{figure}[t]
\hspace{-1cm}
  \includegraphics[width=3.45in]{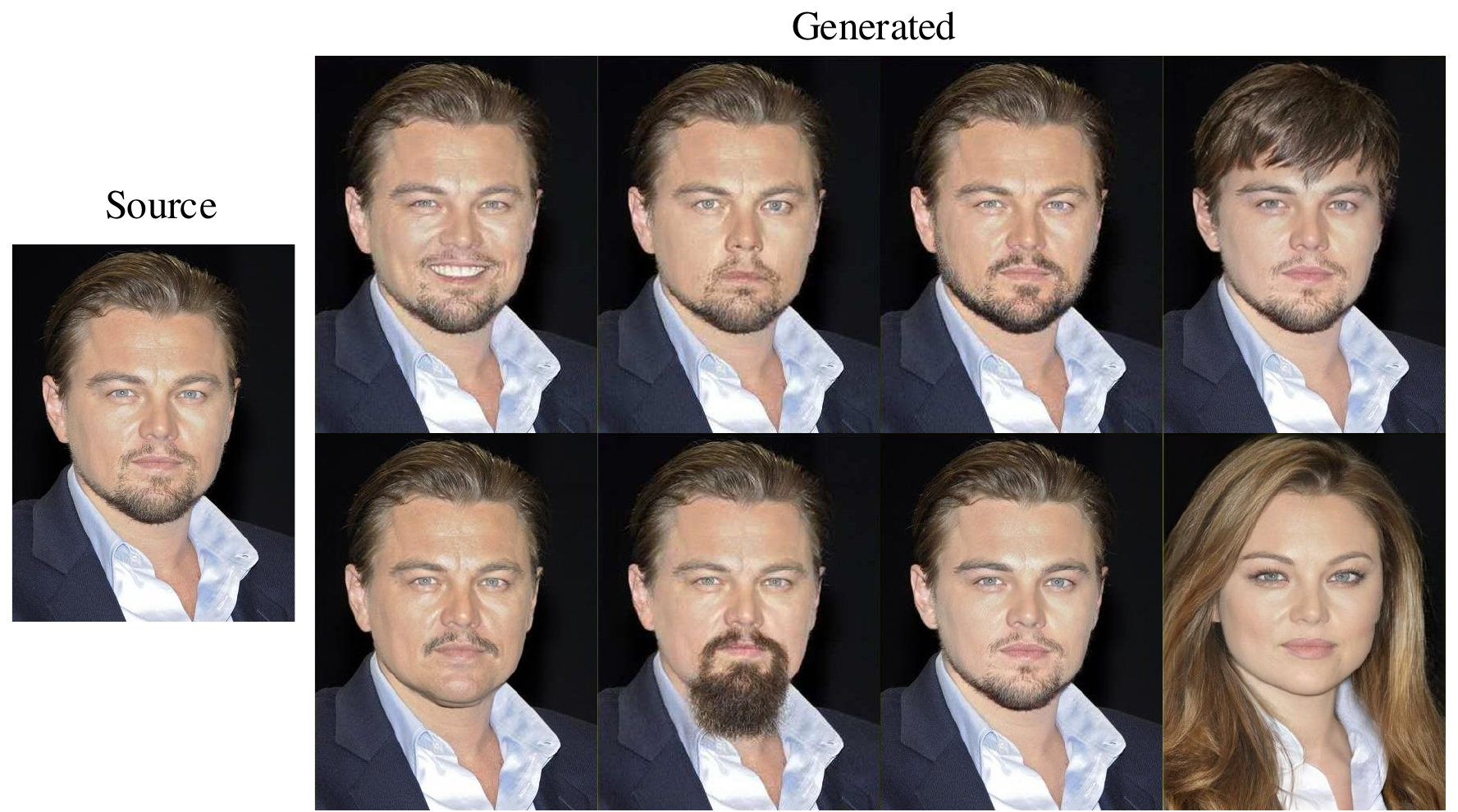}
  \caption{Face style transferring samples with CelebA dataset by \cite{lu2018attribute} \protect \footnotemark.}
\label{fig:s14}
\end{figure}

\begin{figure}[t]
\hspace{-1cm}
  \includegraphics[width=3.5in]{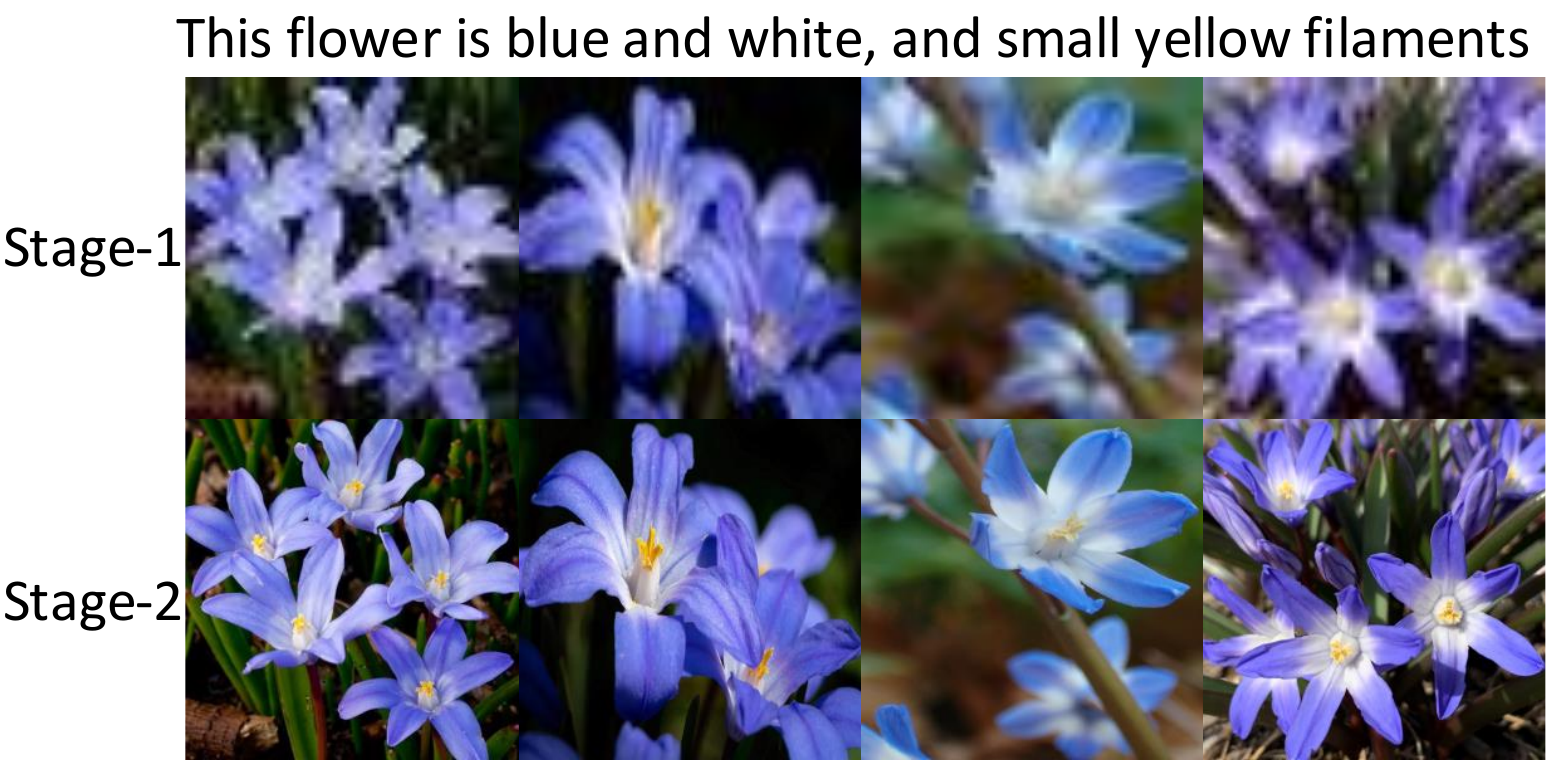}
  \caption{Generated samples of flowers by StackGAN \cite{zhang2017stackgan} \protect \footnotemark.}
\label{fig:s15}
\end{figure}

Figure~\ref{fig:s13} demonstrates the difference between the single-pathway and two-pathway networks and Figure~\ref{fig:s14} contains some samples of face style transferring. We generated these results based the codes provided in \cite{lu2018attribute} on CelebA dataset.
This model uses the attribute image as an identity to create a corresponding conditional vector by incorporating an additional face verification network, for producing high-quality results via a multi-paths network. Choi et al. \cite{choi2018stargan} proposed StarGAN, which is a scalable model that can perform image-to-image translations for multiple domains. The integrated model architecture of StarGAN allows simultaneous training of multiple datasets within different domains of a single network. However, this model requires parallel utterances, transcriptions, or time alignment procedures that decline the efficiency. 
Later, StarGAN-VC \cite{kameoka2018stargan} was proposed to handle the limitations of the preliminary version \cite{choi2018stargan}. StarGAN-VC \cite{kameoka2018stargan} simultaneously learns many-to-many mappings across different attribute domains using a single generator network, it also able to generate converted speech signals quickly enough to allow real-time implementations and only requires several minutes of training to generate reasonably realistic data. 
\footnotetext{{\fontfamily{qcr}\selectfont \color{purple}{https://github.com/bluer555/CR-GAN}}}
Reed et al. \cite{reed2016generative} was the first work that successfully generated plausible images for birds and flowers from their text descriptions. Later, SatckGAN \cite{zhang2017stackgan} is proposed which is the improved version of \cite{reed2016generative}. It has only two layers of the generators. The generator takes $(z, c)$ as input and outputs a blurry image that can present an irregular shape with blurry details of some objects, whereas the second generator as input receives $(z, c)$ and the output of the previous generator and then produces an image with more realistic details. In addition, the proposed StackGAN \cite{zhang2017stackgan} improves upon the preliminary study \cite{reed2016generative} in terms of accuracy, but a look at the structure has a lot to answer, especially about the quantification of speed improvements which is very important in this context. Figure~\ref{fig:s15} presents some generated sample results from StackGAN \cite{zhang2017stackgan}. 
AttenGAN \cite{xu2018attngan} is another latest work that proposed attentional generative adversarial network that uses attention-driven, hierarchy network for creating image out of a text. However, the model suffers from the clarity of the presentation and evaluations. However, the authors claim better results than \cite{reed2016generative} in term of inception scores, nonetheless the results are only empirical and no theoretical analysis is carried out. \\

Doan et al. \cite{doan2019line} introduced a method for training the $G$ against an ensemble of $D$. This problem formalized within the full-information adversarial framework, where the ability of the algorithm is assessed to select combinations of $D_s$ for providing the G with response during learning. 
Hence, a reward function is used which returns the Gs progress and accordingly update the combination weights allotted to each $D$. Pu et al. \cite{pu2019sdf} proposed to fuse disparity maps from different sources, while incorporating additional information (intensity, gradient, etc.) into a refiner network to better refine the raw disparity inputs. 
In \cite{wang2018high}, the authors proposed a model called Photo-Sketch Synthesis by using multi-adversarial networks, iteratively the proposed model generates low resolution to high resolution images in an adversarial way. The hidden layers of the generator are supervised to first create lower resolution images followed by inherent refinement through the network to form more realistic images. 

To produce extremely realistic images, in \cite{wang2018high} the network is normalized by adopting forward backward consistency. This is carried out by introducing cycle consistency losses at different resolution levels, which are formulated as:
\begin{equation}
\begin{aligned}
\ell_{cyc_{Ai}} = \Vert Rec_{Ai} - R_{Ai}\Vert_1 = \Vert G_B(G_A(R_A))_i - R_{Ai}\Vert_1 \\
\ell_{cyc_{Bi}} = \Vert Rec_{Bi} - R_{Bi}\Vert_1 = \Vert G_A(G_B(R_B))_i - R_{Bi}\Vert_1 
\end{aligned}
\end{equation}
where $R_{Ai}$ and $R_{Bi}$ denote the images in different resolutions and $Rec_{Ai}$ and $Rec_{Bi}$ represent the reconstruction outputs. 
In \cite{dong2017musegan}, the authors proposed three approaches for producing music nots by using GANs. The approaches have different architectures and underlying assumptions. Bhattacharjee and Das \cite{bhattacharjee2017temporal} proposed to use two stages of GANs to generate crisp and clear set of the future frames. The main contribution lies in formulating two objective functions based on the normalized cross correlation and the pairwise Contrastive divergence. Although the model is well discussed and all the stages of the model has been evaluated, the overall model's performance was not compared with that of the other state-of-the-art models.
Xiong et al. \cite{xiong2018learning} showed the movement of the clouds with a GAN based two-stage approach to generating realistic time-lapse videos of high resolution. In their model, the first stage generates videos where each frame contents realistic objects. In the second stage, the generated videos of the first stage are improved by enforcing them to be closer to the real video’s frames with regard to motion dynamics. In \cite{aafaq2019video}, the authors collected wide range of datasets and evaluation metrics for video description. Park et al. \cite{park2018mc} proposed a multi-conditional GAN (MC-GAN) which manages both the object and background information equally. The proposed model contains a synthesis block which separates the object and background information during training. This block powers up MC-GAN to generate high realistic images by preserving the background information from the given base images. In \cite{park2018mc}, also the authors based on MC-GAN, introduced a model to produce high resolution images by adding the StackGAN to generator. Olabiyi et al. \cite{olabiyi2018multi} proposed an adversarial network to generate multi-turn dialogue answers. The framework is based on conditional GANs. The generator is a hierarchical recurrent encoder-decoder network and the discriminator is a RNN that shares context and word embedding with the generator.
\\

Azadi et al. \cite{azadi2018multi} proposed to combine few-short learning and a double generator adversarial network for font style transferring that has high computation costs. The study contains a comprehensive review section, however only few of them used to evaluate the performance of the model. Yu et al. \cite{yu2019exploiting} proposed a symmetric approach for heterogeneous image-to-video adaptation, which augments deep images and video features by learning domain-invariant representations of source images and target videos. The model focuses on unsupervised scenarios where the labeled source images are accompanied by unlabeled target videos in the training phrase, also a data-driven approach is presented to learn the augmented features of images and videos with high transform-ability \cite{tabik2020mnist}.
\footnotetext{{\fontfamily{qcr}\selectfont \color{purple}{https://github.com/hanzhanggit/StackGAN}}}
Furthermore, Jang et al. \cite{jang2018video} proposed an Appearance-Motion Conditional GAN to address the challenge of future uncertainty. This model uses motion information as conditions that specify the futures structure and consequently reduce the uncertainty. This model contains one G, two $D_s$ and a ranking module that encourages similar condition videos to look the same. This technique allows to learn different factors of variation and perform different gestures, but it requires massive computational budgets compares to the other methods and also it suffers from motionless and blurry effects that caused by the lack of supervision signals or sub-optimal solutions in the training process. Similar idea can be also found in \cite{zhang2018face} which is  proposed a novel face sketch synthesis method by multi-domain adversarial learning. Unlike \cite{jang2018video}, this model overcomes the defects of blurs, generate high quality synthetic data, and reduces the time consuming.
The principle of scheme relies on the concept of {\itshape interpretation through synthesis}. In particular, the authors interpret face photographs in the photo-domain and face sketches in the sketch domain by reconstructing themselves respectively via adversarial learning. \par To further improve the synthesis speed, the authors in \cite{pang2018visual} introduced a multi modules adversarial network by minimizing a new loss function consisting of pixel-wise loss, adversarial loss and perceptual loss in discriminator for visual haze removal. Although, fine textures could be well generated by using this technique, but deformation and noise still exist in the results.   
\begin{figure}[b]
\centering
  \includegraphics[width=2.3in]{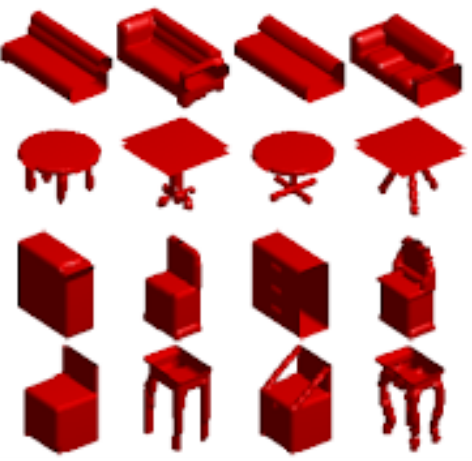}
  \caption{3D completion results on real-world scans. 3D-ED-GAN shows the low-resolution completion result when going through LRCN.}
\label{fig:s17}
\end{figure}
\subsection{Applications to medical imaging:}
In general there are two methods to use GANs in medical imaging. The first one is centralized on the generative phase, which possibly aid in realizing the primary structure of training data to create realistic images. This asset makes GANs very capable in handling data scarcity and patient privacy \cite{esteban2017real, bissoto2018skin, yang2020smile, han2019synthesizing, liu2020tomogan}. The second one is centralized on the discriminative phase, in which the discriminator $D$ can be considered as a learned prior for wild images therefore it can be used as detector for fake generated images \cite{xue2018segan, amirrajab2020xcat}.

\subsubsection{Generative phase}
Sandfort et al. \cite{sandfort2019data} proposed a data augmentation model based on CycleGAN to improve generalizability in CT segmentation. Han et al. \cite{han2020madgan} proposed a two stage unsupervised anomaly detection method for MRI scans based on GAN. In \cite{welander2018generative}, the authors compare the results of two unsupervised GAN models (CycleGAN and UNIT) for image-to-image translation of T1- and T2-weighted MR images, by comparing the created synthetic MR images to real images.
 
\subsubsection{Discriminative phase}
Tang et al. \cite{tang2018ct} proposed a method for CT images segmentation based on stacked generative adversarial networks. The first network layer reduces the noise in the CT image and the second layer creates a higher resolution image with enhanced boundaries. In \cite{schlegl2019f}, the authors proposed a GANs based on unsupervised learning approach which able to identify anomalous images. The proposed model contains fast mapping technique of new data to the GAN's latent space. The mapping is based on a trained encoder. Osokin et al. \cite{osokin2017gans} proposed GANs model to generate the synthesis of cells imaged by fluorescence microscopy. In \cite{yang2018low}, the authors proposed a CT image denoising model based on the GANs with Wasserstein distance and perceptual similarity. Dou et al. \cite{dou2018pnp} proposed GANs for MRI and CT to tackle the significant domain shift by supporting the feature spaces of source and target domains in an unsupervised manner.
%%%%%%%
\subsection{Applications to 3D Reconstruction}
The methods that are listed in this section are used in 3D shape completion. Wang et al. \cite{wang2017shape} proposed a hybrid architecture that used a 3D Encoder-Decoder generative adversarial network with a recurrent convolutional network (LRCN). The 3D-ED-GAN is a 3D network that trained with an adversarial paradigm to fill the missing data in the low-resolution images. Recurrent neural network approach is used to reduce memory usage. Figure~\ref{fig:s17} represents some shape completion examples on real-world scans. 3D-ED-GAN is used to represent the low-resolution output of 3D-ED-GAN and LRCN. 
\begin{figure*}
 \centering
\includegraphics[width=6.5in]{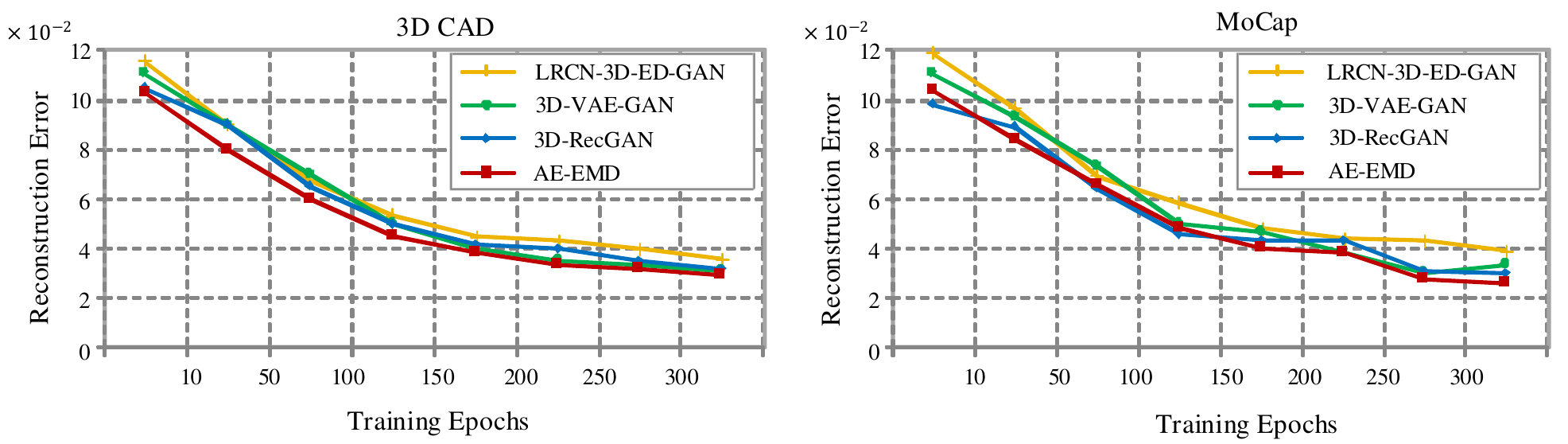}
\caption{Reconstruction error of the various 3D GANs models on 3D CAD and MoCap datasets at various training epochs. Reconstruction measures the closeness of the synthetic results to the training test ground truth distributions.}
\label{fig:s19}
\end{figure*}
Wu et al. \cite{wu2016learning}, proposed 3D Generative Adversarial Network (3D-VAE-GAN), which generates 3D objects from a probabilistic space by leveraging recent advances in volumetric convolutional networks and generative adversarial nets. In their model, the generator launches a mapping from a low-dimensional probabilistic space to the space of 3D objects, to directly reconstruct a 3D object from a 2D input image. The proposed model has a simple structure and the authors clam high reconstruction performance, but the evaluation results and evaluation parameters are limited. In \cite{smith2017improved}, the authors introduced a new model for GAN training to realize the detailed 3D shape of objects. The model adopts Wasserstein normalization with gradient penalization for training which improves realistic image generation. This architecture even can reconstruct 3D shape from 2D images and perform shape completion. 

Yang et al. \cite{yang20173d}, proposed a 3D-RecGAN model, which reconstructs the complete 3D structure of a specified object from one random depth view. Different from the current models which typically needs multiple views of the same object or class labels to recover the full 3D geometry, the proposed model only takes the voxel grid representation of a depth view of the object as input, and is able to produce the complete 3D occupancy grid by filling in the missing regions. Moreover to optimize the generator, a weight $\beta$ is assigned to the reconstruction loss $1 - \beta$ to  $\ell_{GAN}^g$. Therefore, the loss function for $G$ and $D$ defined as follows:
\begin{equation}
\begin{aligned}
\ell_g =  \beta\ell_{rec} + (1 - \beta)\ell_{GAN}^g \\ 
\ell_d = \ell_{GAN}^g
\end{aligned}
\end{equation}
In \cite{galama2019itergans}, the authors proposed an Iterative GANs which iteratively transforms an input image into an output image based on the geometry and appearance of the objects. The model is evaluated on the several dataset, however, the model was not able to generate highly realistic samples.  Hermoza and Sipiran \cite{hermoza20183d} proposed an encoder-decoder 3D deep neural network on the GAN architecture, combining two loss objectives: a completion loss and an Improved Wasserstein GAN loss for 3D object reconstruction.
Achlioptas et al. \cite{achlioptas2018learning} proposed algebraic manipulations and a deep auto-encoder GAN (AE-EMD) for semantic part editing, shape analogies and shape interpolation, as well as shape completion for 3D objects. The reconstruction error of the various 3D GAN models on 3D CAD dataset at different training epochs is shown in Figure~\ref{fig:s19}.

\subsection{Image fusion}
Generating a new image from the set of input images is an interesting research area in the GANs. In \cite{joo2018generating} the authors proposed a GAN-based framework called FusionGAN that generates a fusion image by manipulating two input images. The authors demonstrated that FusionGAN is able to change the input shape and characteristics and generate a new image, while preserving the main content of the inputs. Zhan et al. \cite{zhan2019spatial} proposed a new fusion methods as SF-GAN to synthesize realistic images from the foreground
objects and background images. The authors proved the effectiveness of their model through a comprehensive set of experiments. In addition, several methods have been proposed using GAN's architecture in order to transform an input into a desired shape and improve the fusion performance \cite{ma2020pan, rey2020fucitnet, ma2019fusiongan, zhang2020mff}.

%---------------------------------
\subsection{Image Completion}
Image completion, as a conventional image processing task, intended to fill the missing or masked parts in images that have reasonable synthesized contents. The produced contents are either having much detail as the original one, or easily fit into the image context which appears to be visually realistic. Most of completion methods \cite{chandak2019semantic}are based on low-level cues to look for patches from neighbour regions of the image and create the synthesize contents that are similar to the patches. Unlike the existing models that look for patches to synthesize, the model that proposed by \cite{li2017generative}, produces contents for missing regions based on a CNN. The proposed algorithm is trained with number of reconstruction loss, two adversarial losses and a semantic parsing loss, to ensure pixel quality and local-global contents stability. 
\\

\begin{figure}[t]
%\hspace{-1cm}
\centering
  \includegraphics[width=3.5in]{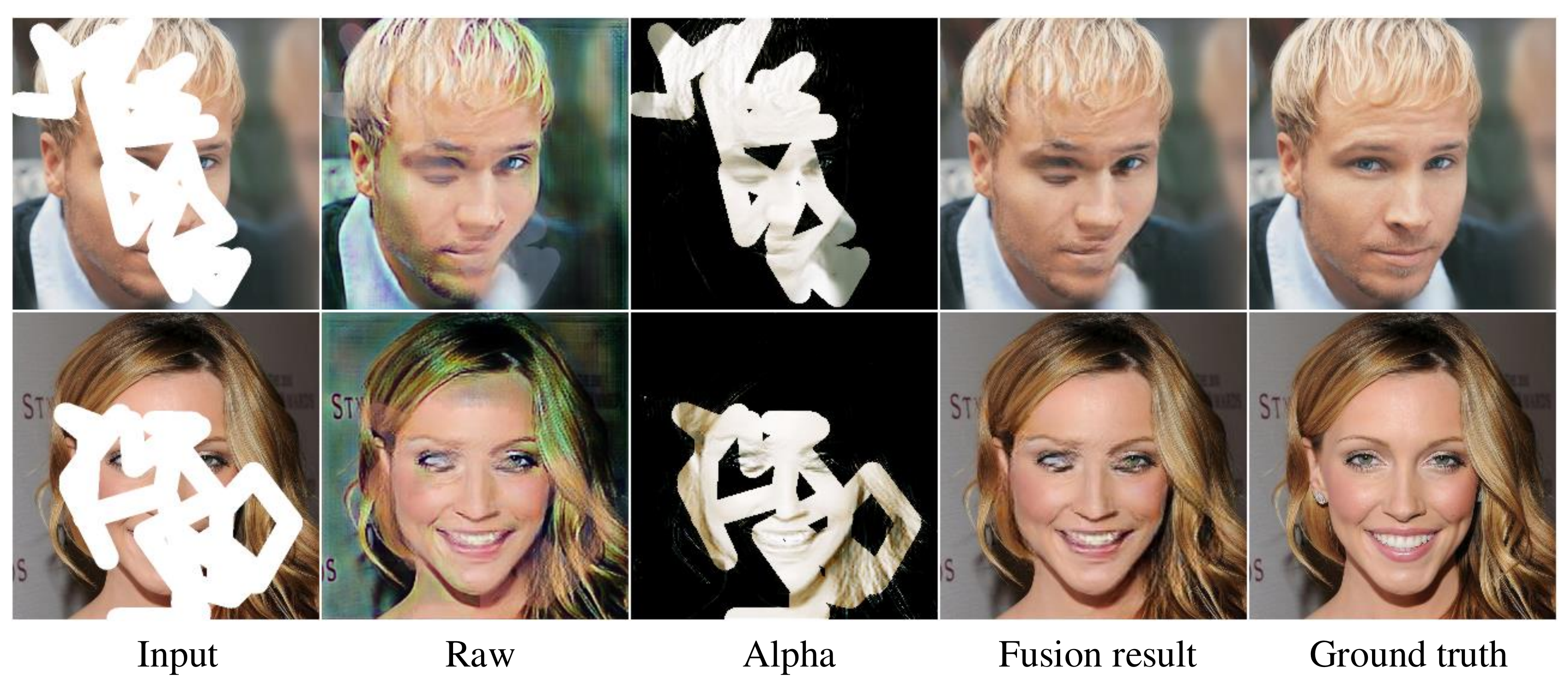}
  \caption{Sample generated results by \cite{hong2019deep}.}
\label{fig:image-completion}
\end{figure}

In \cite{zheng2019pluralistic} the authors proposed a double path framework for image completion. One is a reconstructive path that uses a single ground truth image to get prior distribution of missing parts and recreate the input image from this distribution. The other one is a generative path in which the conditional prior is linked to the distribution gained in the reconstructive path. These methods often fail in human body images that require accurate structure and appearance synthesis. To tackling this problem, \cite{wu2019deep} proposed a double stage method. In which in the first stage, a complete body part generated from an incomplete one through a human parsing network, which closely follows structure recovery within the unknown area through the use of full-body pose approximation. In the second stage, an image completion network is employed to fill the new portions, based on the first stage. 
In \cite{hong2019deep}, to perform image completion a fusion block is introduced to generate a flexible alpha composition map for combining known and unknown regions. 
The fusion block not only provides a smooth fusion between restored and existing content but also provides an attention map to make the network focus more on the unknown pixels. However, the model shows good performance on celebA dataset but cannot perform well on high-resolution images as Figure \ref{fig:image-completion} shows.
\begin{figure}[t]
\centering
  \includegraphics[width=3.5in]{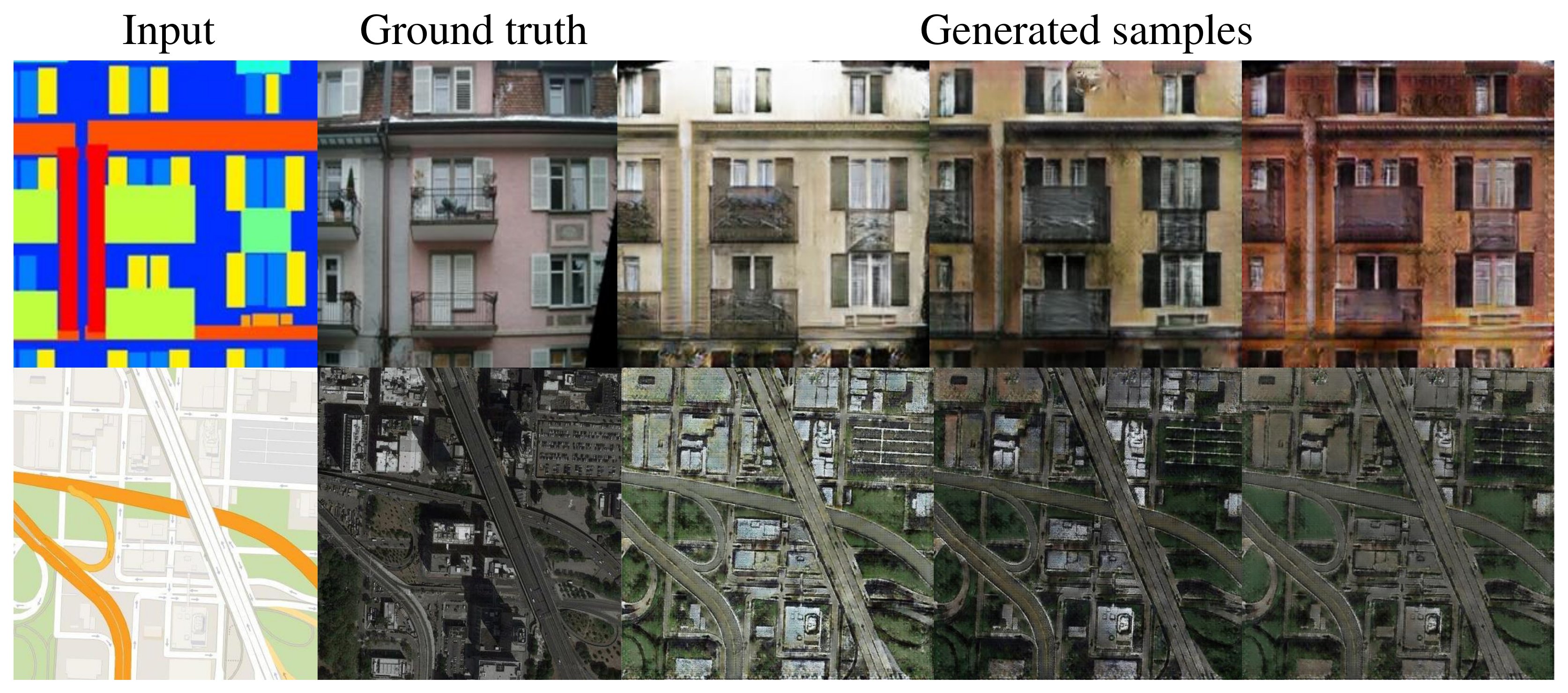}
  \caption{Sample generated results by \cite{isola2017image}.}
\label{fig:g-43}
\end{figure} 
%%%%%%%%%%%
 
\section{Image-to-Image-Translation}
\label{sec:6}
Most of computer visions problems can be seen as an image-to-image translation problem, mapping an image from one domain to another image in different domain. As an illustration, super-resolution can be viewed as a concern of mapping a low-resolution image to a similar high-resolution one; image colorization is a problem of mapping a gray-scale image to a corresponding color one. The problem can be investigated in supervised and unsupervised learning methods. In the supervised approaches, paired of images in various domains are available \cite{isola2017image}. 
In the unsupervised models, only two separated sets of images are available in which one composed of images in one domain and the other composed of different domain images—there is no paired samples representing how an image can possibly translated to a corresponding image in different domain. For lack of corresponding images, the unsupervised image-to-image translation problem is considered more difficult, but it is more feasible because training data collection is easier. 

When assessing the image translation problem from a likelihood viewpoint, the main challenge is to learn a mutual distribution of images in different domains. In the unsupervised setting, the two sets composed of images from two minor distributions of different domains, and the task is to gather the cooperative distribution by utilizing these images. However, driving the joint distribution from the minor distributions is extremely ill-posed problem \cite{ma2018disentangled}.
In this section, we discuss the image-to-image translation methods. Image-to-image translation is similar to style transfer \cite{jing2019neural}, which as the input receives a style image and a content image. The model output is an image that has the content of the content image and the style of the style image. It is not only transferring the images' styles, but also manipulates features of objects. This section lists several models that are proposed for image-to-image translation from supervised methods to unsupervised ones. Figure \ref{fig:g-43} shows sample generate results by \cite{isola2017image}. 
%%%%%%%%%

\subsection{Supervised Translation}
Isola et al. \cite{isola2017image} proposed to merge the different network losses of Adversarial Network with $L_1$ regularization loss, therefore the particular generator not only trained to pass the discriminator filtering but also to produce images that contain realistic objects and similar to the ground-truth images. $L_1$ generates less blurry images as compared to $L_2$, it was the reason for using $L_1$. The conditional GAN loss is formulated as:
\begin{equation}
\begin{aligned}
\ell_{cGAN} (G,D) = E_{(x,y) \sim p_{data} (x,y)}[\log D(x,y)] + \\
E_{x \sim p_{data} (x), z \sim p_z (z)} [\log (1 - D(x, G(x, z))]. 
\end{aligned}
\end{equation}
in which $x, y \sim p(x, y)$ denotes to the images that have different styles but belong to the same scene, similar to the standard GAN \cite{goodfellow2014generative}, $z \sim p(z)$ represents random noise, thereby $L_1$ loss for pressuring self-similarity is defined as:
\begin{equation}
\ell_{L_1} (G) = E_{x,y\sim p_{data}(x,y)} , z\sim p_z(z) , [||y - G(x, z)||_1],
\end{equation}
the general objective is specified by: 
\begin{equation}
G^* , D^* = arg^{\min_G \max_D} \ell_{cGAN}(G,D) + \lambda \ell_{L_1}(G)
\end{equation}
in which the hyperparameter of $\lambda$ is used to balance the two loss functions. Moreover, in \cite{isola2017image}, the authors pointed out that, the noise $z$ does not have noticeable influence on the result, therefore, they proposed to use the noise in the form of dropout during training and test in place of samples that belongs to random distribution.
In this model, the structure of the $G$ is based on the new structure of U-Net \cite{shamsolmoali2019novel} that has multi-scale connections to join each encoder layer to the same layer decoder for sharing low-level information like edges of objects. In \cite{isola2017image} the authors proposed {\verb|PatchGAN|}. The proposed model rather than classifying the whole image attempts to classify the {\verb|N x N|} path of each image and seek the average scores of patches for obtaining the final score of the image. From the experiments it has been observed, for obtaining the high frequency details, it is sufficient to limit the discriminator to focus on the local patches. 
\begin{figure}[t]
\centering
  \includegraphics[width=2.9in]{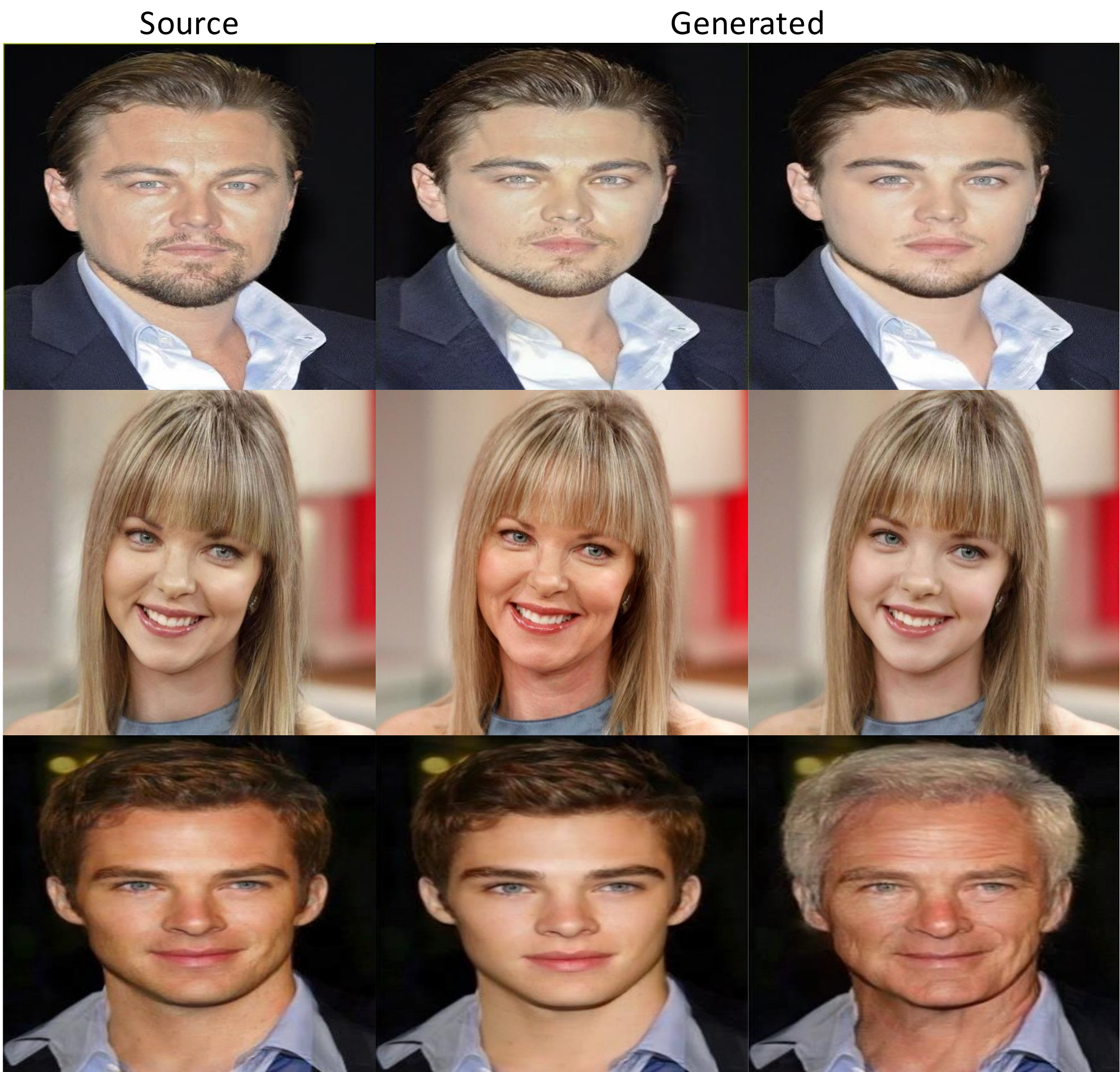}
  \caption{Comparison of generated results and ground truth for face aging on CelebA datasets by using \cite{wang2018face}.}
\label{fig:s21}
\end{figure}

Yoo et al. \cite{yoo2016pixel} proposed an algorithm for supervised image-to-image translation, while having a secondary discriminator $D_{pair}$ that evaluates whether or not a pair of images from multiple domains is related with each other. The loss of $D_{pair}$ is calculated as follows:
\begin{equation}
\begin{aligned}
\ell_{pair} = - t\log [D_{pair}(X_s, X)] \\
+ (t - 1)\log [1 - D_{pair}(X_s, X)], \\
s.t.t =
\begin{cases}
0 & {if X = X_t} \\  
0 & {if X = \hat X_t} \\
0 & {if X = X_{\overline t}}
\end{cases}
\end{aligned}
\end{equation}
where the input image from the source domain is represented by $X_s$ and its groundtruth image is denoted by $X_t$ in the target domain, an irrelevant image in the target domain is represented by $X_{\overline t}$. The generator in the proposed model transfers $X_s$ into a single image $\hat X_t$ in the associated domain. 
In \cite{shamsolmoali2020road}, the authors proposed an efficient pyramid adversarial networks to generating synthetic labels based on target domains for road segmentation in remote sensing images. Zareapoor et al. \cite{zareapoor2020oversampling} proposed a semi-supervised adversarial networks for dataset balancing in mechanical devices. 
In \cite{shamsolmoali2020amil}, the authors integrate multi-instance learning into adversarial networks for human pose estimation. As the results show, the proposed model has high accuracy and fast performance. Shamsolmoali et al. \cite{shamsolmoali2020imbalanced} to handle the imbalanced class problems, proposed a capsule adversarial networks based on minority class augmentation.

 In \cite{gonzalez2020decomposition}, the authors proposed a general learning framework assign the generated samples to a distribution over a set of labels instead of a single label. The effectiveness of their proposed model is proved through a set of experiments.  Zhang et al. \cite{zhang2018drcw} proposed DRCW-ASEG method in order to generate synthetic examples for multi-class imbalanced problem. The authors shown that their proposed strategy is able to improve the classification accuracy.  

\subsection{Unsupervised translation}
Zhu et al. \cite{zhu2017unpaired} and Yi et al. \cite{yi2017dualgan} proposed to adopt a reconstruction loss to maintain the input image quality after a cycle of alteration. These approaches share the similar architecture as shown in Figure~\ref{fig:s6}(b). The two generators $G_{AB}$ and $G_{BA}$ are performing reverse transformations and gets benefits from dual learning \cite{he2016dual}. In addition, another model is proposed by Kim et al. \cite{kim2017learning} that adopted a similar cyclic architecture. In the proposed architecture, two generators are used $G_{AB}$ to transfer an image from domain $A$ to $B$ and $G_{BA}$ is used to sends the image back to the original domain. 
\begin{figure}[t]
  \includegraphics[width=3.3in]{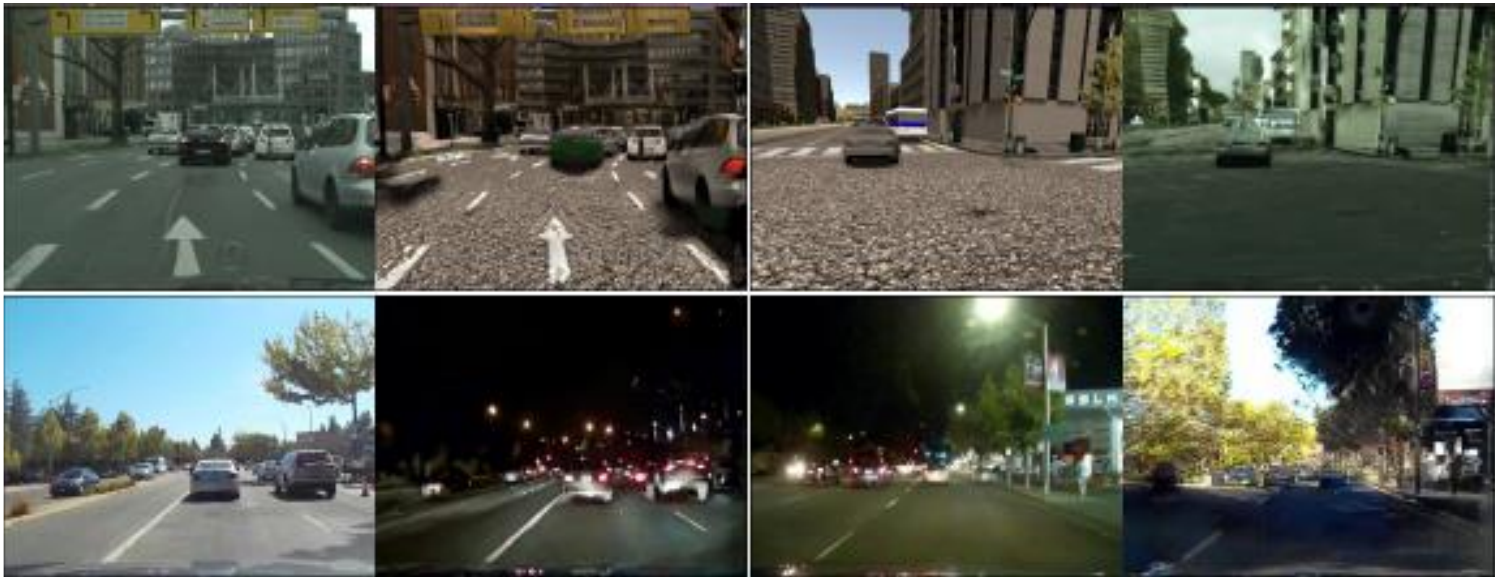}
  \caption{Street scene image translation sample results by \cite{liu2017unsupervised}. For each pair, left is input and right is the translated image.}
\label{fig:s22}
\end{figure}
%%%%%%
\begin{figure}[t]
  \includegraphics[width=3.3in]{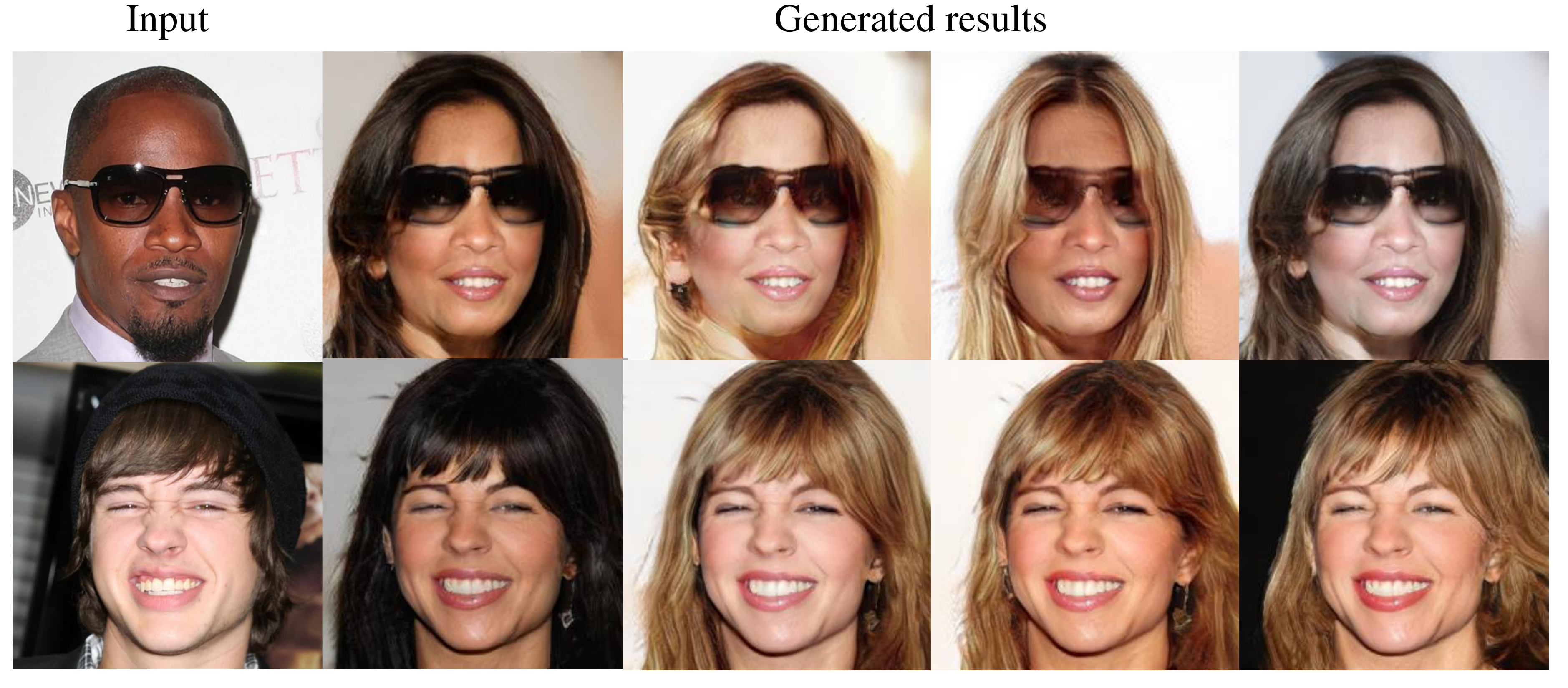}
  \caption{Male to female image translation sample results by \cite{liu2016coupled}. For each row, left is input and right is the translated samples.}
\label{fig:men to female}
\end{figure} 
\begin{figure}[t]
  \includegraphics[width=3.3in]{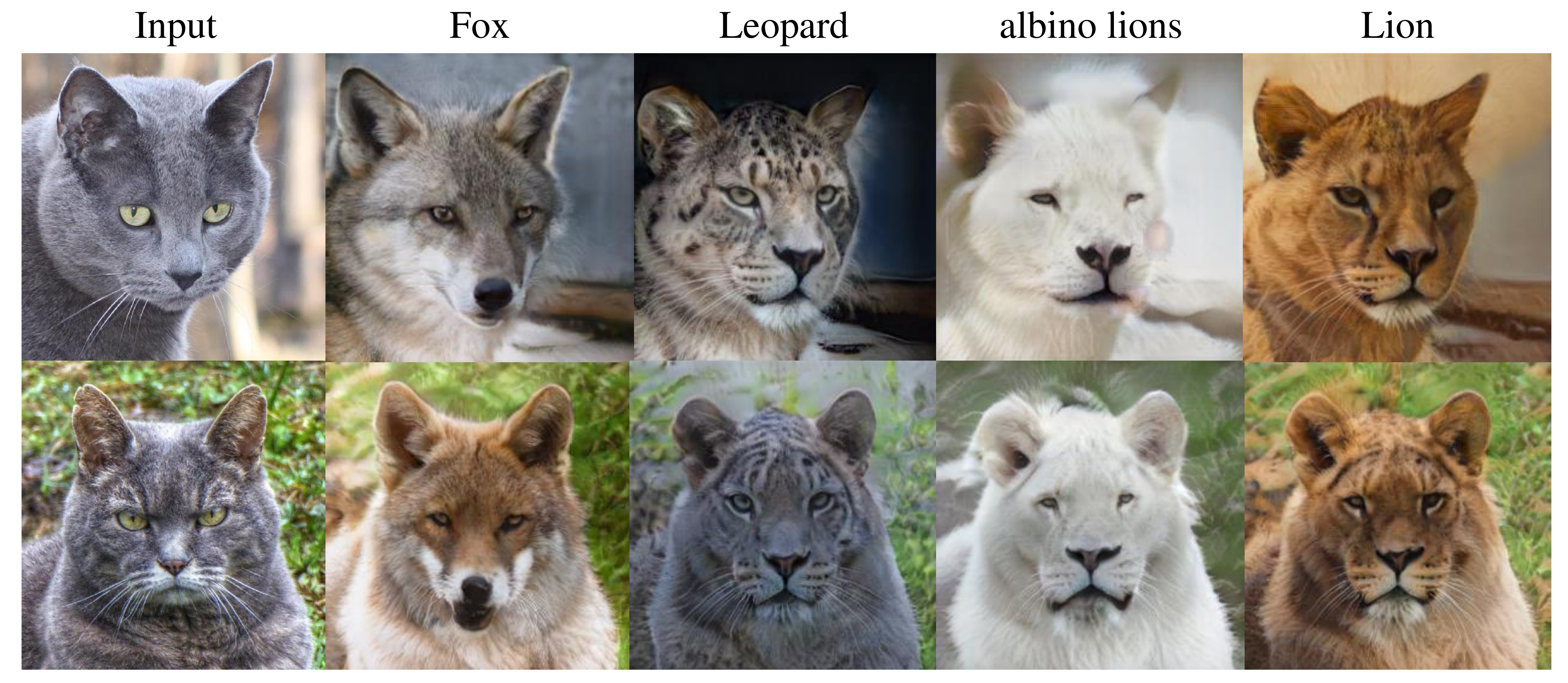}
  \caption{Cat to other species translation results by \cite{liu2017unsupervised}.}
\label{fig:cat to wild}
\end{figure}
There are also two discriminators $D_A$ and $D_B$ to distinguish the domain of the image. The adversarial loss function for $G_{AB}$ and $D_B$ is defined as:
\begin{equation}
\begin{aligned}
\ell_{GAN}(G_{AB},D_B) = E_{b\sim p_{B(b)}}[\log D_B (b)] \\
+ E_{a\sim p_{A(a)}}[1 - \log D_B(G_{AB}(a))]
\label{eq:e23}
\end{aligned}
\end{equation}
for $G_{BA}$ and $D_A$ the adversarial loss is defined $\ell_{GAN}(G_{BA}, D_A)$ in which the data distribution is defined as $b\sim p_{B(b)}$ and $a\sim p_{A(a)}$ in different domains. The adversarial loss is aimed to reduce the reconstruction error once translating an image from one domain to another one and later translating it back to the original domain,\\ i.e.,$a \rightarrow G_{AB}(a) \rightarrow G_{BA}(G_{AB}(a)) \approx a$ The cycle-consistency loss is determined as:
\begin{equation}
\begin{aligned}
\ell_{cyc}(G_{AB}, G_{BA}) = E_{a\sim p_A(a)}[||a - G_{BA}(G_{AB}(a)||_1] \\
+ E_{b\sim p_B(b)}[||b - G_{AB}(G_{BA}(b))||_1]
\end{aligned}
\label{eq:e24}
\end{equation}
consequently, the overall loss function is defined as:
\begin{equation}
\begin{aligned}
\ell(G_{AB}, G_{BA}, D_A, D_B) = \ell_{GAN}(G_{AB}, D_B) \\
+ \ell_{GAN}(G_{BA}, D_A) \\
+ \lambda \ell_{cyc}(G_{AB}, G_{BA})
\end{aligned}
\end{equation}
in which $\lambda$ denotes a hyperparameter for balancing the losses and the objective function is defined as:
\begin{equation}
G_{AB}^* , G_{BA}^* = arg_{G_{AB}}^{\min_{G_{BA}} \max_{D_B, D_A}}loss(G_{AB}, G_{BA}, D_A, D_B)
\end{equation}
%%%%%%%%%%%%%%
The architecture proposed in \cite{zhu2017unpaired, yi2017dualgan} follows the U-Net structure \cite{shamsolmoali2019novel}. In \cite{zhu2017unpaired} the authors stabilized two methods for training. Firstly, rather than using the {\it log loss} \cite{mao2017least} for $\ell_{GAN}$ in Eq.(\ref{eq:e23}) by using {\it least square loss} the model performs more stably and generates higher realistic images:
\begin{equation}
\begin{aligned}
\ell_{LSGAN}(G_{AB},D_B) = E_b\sim p_{B(b)}[(D_B(b) - 1)^2] \\
+ E_a\sim p_{A(a)}[D_B(G_{AB}(a))^2]
\end{aligned}
\end{equation} 
Secondly, to reduce the model oscillation, the discriminators are update $D_A$ and $D_B$ by using the previous generated images. 
The proposed model has significant performance and the results are widely evaluated against those of \cite{isola2017image, zhu2017toward}. In \cite{liu2017face} and \cite{wang2018face}, the authors proposed contextual generative adversarial networks for face aging \footnote{{\fontfamily{qcr}\selectfont \color{purple}{https://github.com/dawei6875797/Face-Aging-with-Identity-Preserved-Conditional-Generative-Adversarial-Networks}}}. 
The model contains a conditional transformation network \cite{engel2017latent} and two discriminative networks. The conditional transformation network handles the aging procedure with a number of residual blocks and the discriminator guide the synthesized face to have real distribution. Some generated sample results are shown in Figure~\ref{fig:s21}.

Benaim and Wolf \cite{benaim2017one} found out that, the distance $\Vert X_i - X_j\Vert$ between the images in the source domain $A$ is extremely dependent on their counterparts distance $\Vert G_{AB}(X_i)- G_{AB}(X_j)\Vert$ in the target domain $B$. If $d_k$ is the distance $\Vert X_i - X_j\Vert$ and $\vert \acute d_k\vert$ high dependency signifies that $\Sigma d_k \acute d_k$ accordingly will be high. 
 \begin{table*}[t]
\centering
\caption{Briefly Discusses Different GAN-Based Methods. From Left to Right, Methods: it's input format, it's output format, it's characteristic, the loss functions, and the backbone code. In the column of loss fucntion, $L_{ADV}$, $L_1$,$L_2$, $L_{SEG}$, $L_{IP}$, $L_{SYM}$, $L_{CYC}$, $L_{DUAL}$, $L_{CLS}$, $D_{KL}$, $L_{latent}$, $L_{WGAN}$, $L_{ID}$, $L_{cyc}^{roi}$, $L_{STYLE}$, $L_{CONTENT}$, $L_R$, $L_T$, $L_S$, denote as; Adversarial Loss, $L_1$ distance,$L_2$ distance, Segmentation Loss, Identity Preserving Loss, Symmetry Loss, Cycle Consistency Loss, Dual Learning Loss, Classification Loss, Kl Divergence, Latent Consistency Loss, Cycle-Consistent Loss, Pixel-Wise Identity Mapping Loss, Focal Loss, Style Transfer Loss, Content Loss, Loss Reconstruction, Task Loss, Smooth Loss, Semantic Consistency Loss, Geometry Consistency Constraint Loss; respectively. In The Column Of Code, T, TF, and PT, refer to Torch, Tensorflow, Pytorch, Respectively.}
\label{table:nooonlin}
\scriptsize
% \resizebox{\textwidth}{!}{%
%\hspace{-12mm}
\begin{tabular}{p{.12\textwidth}p{.15\textwidth}p{.3\textwidth}p{.22\textwidth}p{.03\textwidth}}
\hline
Model & Input $\sim$ Output & Characteristics & Loss Function & Code\\  [2ex]
\hline 
BiGAN \cite{donahue2016adversarial} & wild image$\sim$wild image & A supervised method for feature learning & $L_{adv}$ & TF \\ [0.5ex]
ALI \cite{dumoulin2016adversarially} &face$\sim$face & An unsupervised model which jointly learns
a generation network and an inference network & $L_{GAN}$ & TF \\  [0.5ex]
StarGAN \cite{choi2018stargan} &face$\sim$face & An unsupervised model for multiple domains image translations & $L_{adv}+L_{rec}$ & PT+TF \\  [0.5ex]
BicycleGAN \cite{zhu2017toward} & single$\sim$multiple & An unsupervised model for many-to-one mapping & $L_1+ L_{adv}+ D_{KL}+ L_{perseptual}$ & PT \\  [3.5ex]
CGAN \cite{lin2018conditional} & face$\sim$face & An unsupervised model for conditional based image translation & $L_{GAN}+L_{dual}$ & TF \\  [0.5ex]
ComboGAN \cite{anoosheh2018combogan} & face$\sim$face & An unsupervised model for multi-domain multi-component image translation & $L_{cycle}+L_{GAN}$ & PT+TF \\  [0.5ex]
CSGAN \cite{yang2017improving}& text$\sim$text & A supervised model for conditional sequence generation&$L_1$& TF \\  [0.5ex]
DRPAN \cite{wang2018discriminative}& image$\sim$paint & A Discriminative unsupervised Region Networks&$L_G+L_D+L_1+L_2+L_{DP}$&PT \\  [3.5ex]
IA-CycleGAN \cite{huang2017face}& face$\sim$face  & An unsupervised identity preservation model for  image-to-video/video-to-image translation&$L_2+L_{GAN}+L_{WGAN}+L_{id}$&TF \\  [0.5ex]
In2i \cite{perera2018in2i}& wild image$\sim$wild image &An unsupervised model for Multi-Image/Multi- modalities translations&$L_{GAN}+L_{latent}$&TF \\  [0.5ex]
IR2VI \cite{liu2018ir2vi}& wild image$\sim$wild image & An unsupervised model for thermal-to-visible image translation&$L_{cycle}+L_{GAN}+(L_{cyc}^{roi})$&PT+TF \\  [0.5ex]
MedGAN \cite{armanious2020medgan}& medical images$\sim$medical images &An unsupervised model that penalizes the discrepancy between the translated images and the desired ones&$L_1+ L_{GAN}+ L_{style}+ L_{perseptual}+L_{content}$ & TF \\  [3.5ex]
Sem-GAN \cite{cherian2019sem}&wild image$\sim$wild image& An unsupervised framework for GAN consistency&$L_{cycle}+ L_{seg}+ L_{seg}^g+L_{dis}$ & PT \\  [3.5ex]
T2net \cite{zheng2018t2net}&wild image$\sim$wild image& An unsupervised widespectrum translator&$L_{GAN}+ L_r+ L_t+L_s$ & PT+T \\  [0.5ex]
TC-GAN \cite{engelhardt2018improving}&video$\sim$image & An unsupervised method with temporal consistency for removing flickering between video frames&$L_{GAN}$&TF+PT \\  [0.5ex]
Twin-GAN \cite{li2018twin}&face$\sim$carton& An unsupervised multi-loss function framework for  one-to-one image mapping&$L_{adv}+ L_{cycle}+ L_{sem}$ & PT\\  [3.5ex]
U-GAN \cite{dong2017unsupervised}&face$\sim$face&Unsupervised dual networks learning& $L_{GAN}$&TF\\  [0.5ex]
Da-GAN \cite{ma2018gan}&wild image$\sim$wild image&Unsupervised attention GAN for instance image translation&$L_{cst}+ L_{sym}+ L_{gan}$&PT+TF \\  [3.5ex]
XGAN \cite{royer2020xgan}& face$\sim$carton & Unsupervised image-to-image translation with a dual adversarial auto-encoder&$L_{rec}+ L_{dann}+ L_{sem}+L_{GAN}+L_{teach}$ &TF \\  [0.5ex]
CCD-GAN \cite{gomez2018unsupervised}&text$\sim$text& An unsupervised model for improving CycleGAN stability&$L_{cycle}+ L_{GAN}$&TF \\  [3.5ex]
GcGAN \cite{fu2019geometry}& wild image$\sim$wild image & An unsupervised model for geometry-consistent domain mapping&$L_{cycle}+ L_{dis}+ L_{GAN}+L_{GEO}$&PT  \\  [0.5ex]
GM-GAN \cite{ben2018gaussian}&face$\sim$face &A supervised Gaussian Mixture GAN for conditional sampling&$L_{GAN}+ L_{adv}$&TF \\  [0.5ex]
GANVO \cite{almalioglu2019ganvo}&video$\sim$image& An unsupervised model for depth map estimation& $L_{GAN}+ L_{adv}$&TF \\  [0.5ex]
VoS-GAN \cite{spampinato2019adversarial}&video$\sim$image& An unsupervised learning model by using self-supervision mechanism&$L_{adv}$&PT+TF \\   [0.5ex]
WaterGAN \cite{li2017watergan}&video$\sim$image&Unsupervised depth pairings for color correction&$L_{adv}$&TF \\   [0.5ex]
\hline
\end{tabular}
\end{table*}
The distances in the source domain $d_k$ are constant and the extending $\Sigma d_k \acute d_k$ leads to a large value for the loss, and in order to reduce it, the authors proposed $\Sigma \Vert d_k - \acute d_k\Vert$. In \cite{yang2020smile} the authors proposed to apply a pairwise distance loss:
\begin{equation}
\begin{aligned}
\ell_{dist}(G_{AB}, G_{PA}) = E_{x_i, x_j\sim p_A}\vert \frac {1}{\sigma_A}(\Vert X_i - X_j\Vert 1 - \mu_A ) \\
- \frac{1}{\sigma_B}(\Vert G_{AB}(X_i) - G_{AB}(X_j)\Vert 1- \mu_B) \vert
\end{aligned}
\end{equation}
while $\mu_A, \mu_B(\sigma_A, \sigma_B)$ are the deviations of the distances in the pre-training sets of domains $A$ and $B$. To maintain the gradient descent where only single sample at a time is fed into the model, another self-distance constriction is proposed as follows:
\begin{equation}
\begin{aligned}
\ell_{{self}-{dist}}(G_{AB}, G_{pA}) = E_{x\sim p_A} \vert \frac{1}{\sigma_A}(\Vert L(X) - R(X) \Vert 1 - \mu_A) \\
- \frac{1}{\sigma_B}(\Vert G_{AB} L(X) - G_{AB}(R(X))\Vert 1 - \mu_B\vert
\end{aligned}
\end{equation}
while $L(x)$ and $R(X)$ represent the left and right sides of the image $X$ in this method, only the right and left sides of the images participated for calculating $\mu_A , \mu_B$. Accordingly, the general loss is calculated by:
\begin{equation}
\begin{aligned}
\ell = a_{1A}\ell_{GAN}(G_{AB}, D_B) + \alpha_{1B}\ell_{GAN}(G_{BA}, D_A) \\
+ \alpha_{2A}\ell_{dist}(G_{AB}, p_A) + \alpha_{2B}\ell_{dist}(G_{BA}, p_B) \\
+ \alpha_{3A}\ell_{self - dist}(G_{AB}, p_A) + \alpha_{3B}\ell{self - dist}(G_{BA},p_B) \\
+ \alpha_4 \ell_{cyc}(G_{AB},G_{BA})
\end{aligned}
\end{equation}
in which $\ell_{cys}(G_{AB}, G_{BA})$ are respectively formulated in Eq.(\ref{eq:e24}) and Eq.(\ref{eq:e23}) for both the domains.
As earlier discussed, in addition to reducing the reconstruction error at ground truth pixel level, it is also possible to perform the process at higher feature levels \cite{taigman2016unsupervised}. The authors proposed to implement two neural networks in the $G$, a $CNN_f$ and a {\it transposed} $CNN_g$ in which $G = (g\in f)$. In this model, $f$ performs as a feature extractor and the network tries to keep high level features of any input image. Here $X_s$ represents the source domain and $X_t$ denotes the target domain. For the input image $x\in X_s$ the $G$ result is $G(x) = g(f(x))$. Consequently, the reconstruction error with a distance $d$ can be defined as:
\begin{equation}
\ell_{CONST} = \sum\limits_{x\in X_s}d(f(x, f(g(f(x)))))
\end{equation}
%%%%%%%
%%%%%%%% 
Liu et al. \cite{liu2017unsupervised, liu2016coupled} proposed an architecture for unsupervised image-to-image translation. In the proposed architecture, there are two encoders, generators and discriminators, in which both the encoders share the same latent space. The model applies weight sharing in the last layers of the encoders and it successfully learns the joint distribution without any tuple of corresponding images. The applications also go to domain adaptation and image transformation. 
In Figure~\ref{fig:s22}, we showed some sample results achieved by \cite{liu2017unsupervised} that translates images between the synthetic images in the {\it SYNTHIA} dataset and the real images in the Cityscape dataset. Figure ~\ref{fig:men to female} shows some sample male to female  translation results by \cite{liu2016coupled}. In \cite{li2018twin}, the author proposed a progressively growing encoder-generator for translating unlabeled images from one domain into analog images in another domain. The model is used for translating face images, without supervised one-to-one image mapping.\\

\begin{figure}[t]
\centering
  \includegraphics[width=2.8in]{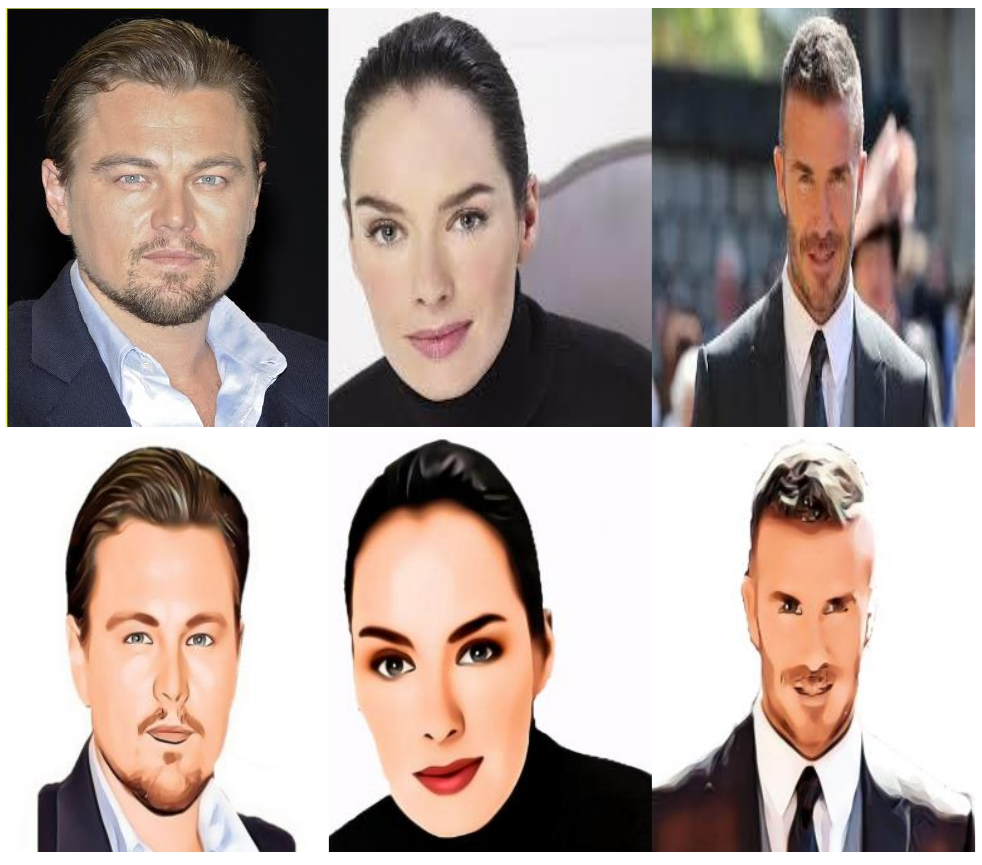}
  \caption{Samples of face to cartoon conversion using XGAN \cite{royer2020xgan} on CelebA dataset.}
\label{fig:s23}
\end{figure}

\begin{figure*}
\hspace{-1cm}
  \includegraphics[width=7.2in]{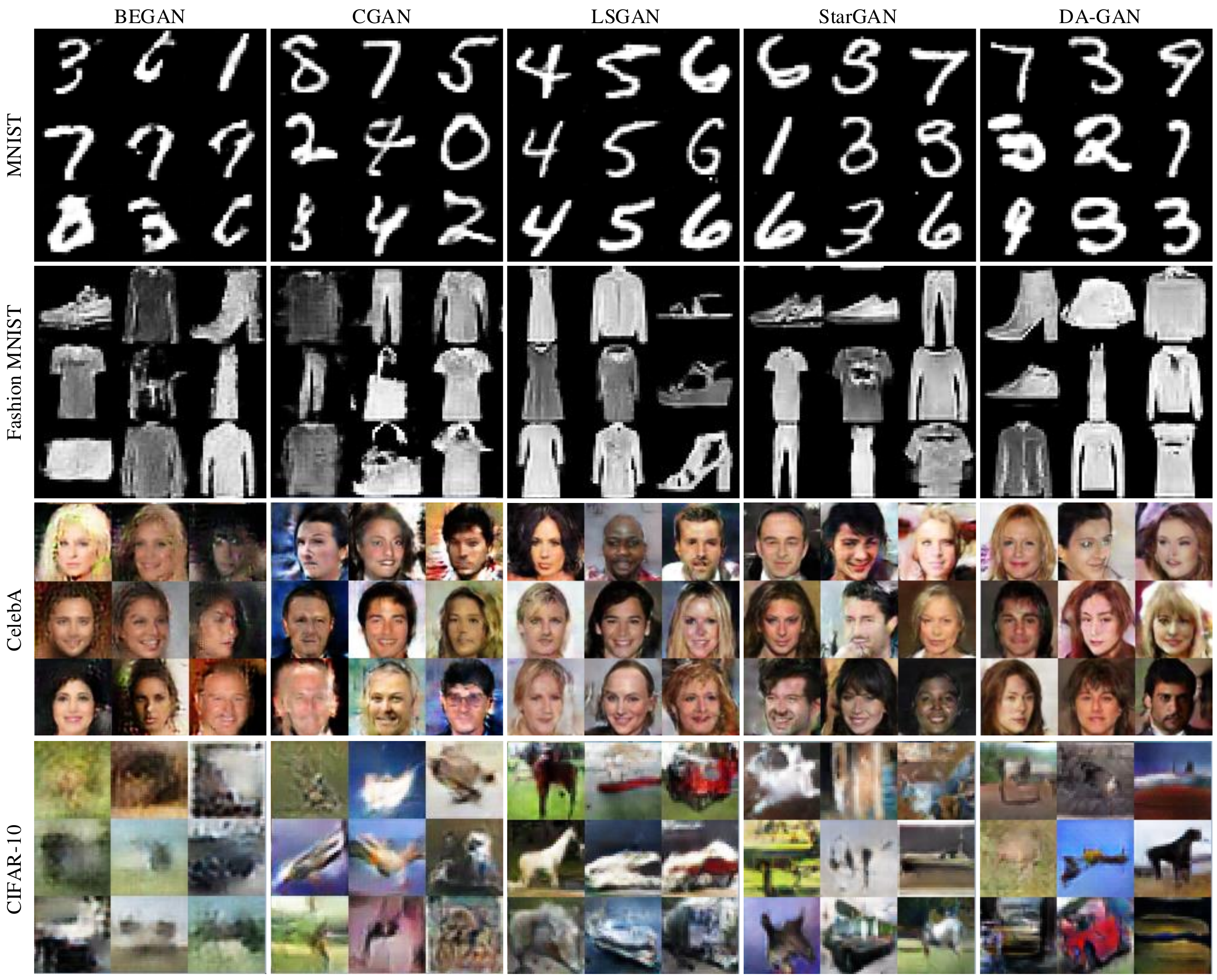}
  \caption{Sample generated results by BEGAN, CGAN, LSGAN, StarGAN, and DA-GAN on four image datasets.}
\label{fig:s16}
\end{figure*}

Furthermore, Royer et al. \cite{royer2020xgan} introduced XGAN architecture that consists of a double adversarial auto-encoder to captures a shared representation of the domain semantic content in an unsupervised manner, by jointly learning the domain-to-domain image translations in both directions. This model reported promising qualitative results for the task of face-to-cartoon translation. 
Figure~\ref{fig:s23} shows some sample results of XGAN. The current approaches only can transform images between two domains, however if we want to transform an image between several domains, a separate generator should be trained on each domain, which has high computation costs. To deal with this problem, Choi et al. \cite{choi2018stargan} suggested using a generator that has the ability of generating the images of different domains. The authors proposed a network that can perform image-to-image translations for multiple domains. The architecture handles simultaneous training of multiple datasets with different domains in a single network. The model has superior performance on a facial attribute transfer and a facial expression synthesis tasks. 
Table~\ref{table:nooonlin} summarizes the input, output, characteristics, learning procedure and implementation platform of several models presented in this review. 
Mordido et al. \cite{mordido2018dropout} proposed to incorporate adversarial dropout in GAN, by dropping out, the feedback of the discriminator in the framework with some probability at the end of each batch. The proposed model releases the generator not to constrain its output to satisfy a single discriminator, but, instead, to fulfill a dynamic ensemble of discriminators. The model has high convergence and the model tested on several dataset, but the authors did not discuss the efficiency of the model. Figure \ref{fig:cat to wild} shows some sample translation results by \cite{liu2017unsupervised}.
%%%%%
 In Figure~\ref{fig:s16} we evaluated the performance of BEGAN \cite{berthelot2017began}, CGAN \cite{hong2018conditional}, LSGAN \cite{mao2017least}, StarGAN\cite{kameoka2018stargan}, and DA-GAN \cite{ma2018gan} on four datasets.

\section{Conclusion and Discussion}
\label{sec:7}
This paper reviewed the existing GAN-variants for synthetic image generation based on architecture, performance, and stable training. We also reviewed the current GAN-related research architecture, loss functions, and datasets that are generally used for synthetic image generation. In particular, it is difficult, yet important for image synthesis tasks to explicitly define the loss. For instance, to perform style transfer, it is difficult to set a loss function to evaluate the matching of an image to a certain style. Each input image in synthetic image generation may have several legitimate outputs, however these outputs may not cover all the conditions. For synthetic image generation, several recent supervised and unsupervised methods have been reviewed, their strengths and weaknesses are thoroughly discussed.

Although we have conducted several experimental evaluation, GANs for synthetic image generation, still lacks a thorough study of domain adaptation and transfer learning. In addition, the computer vision community would benefit from an extension of this practical study that compares in addition to accuracy, the training and testing time of these models. Moreover, we think that the effect of normalization models on the learning capabilities of CNNs should also be thoroughly explored. At the time of this writing, there are a few published works on using GANs for video, time series generation, and natural language processing. Future research should be directed towards investigating the use of GANs in those fields as well as others. 

\bibliographystyle{IEEEtran}
\bibliography{IEEEabrv,Ref}
%\bibliography{Ref}

%\section{REFERENCES}
%\footnotesize{
%\begin{hangparas}{0.2in}{1}
%\noindent [185] ~ Denton, Emily L. and Chintala, Soumith and Fergus, Rob and others,  2018. Deep generative image models using a  laplacian pyramid of adversarial networks. An introduction to image synthesis with generative adversarial nets, Advances in neural information processing systems. pp.1486-1494.
%\end{hangparas}
%\begin{hangparas}{0.2in}{1}
%\noindent [186] L. Metz, P. Ben, P. David, and S. Jascha. "Unrolled generative adversarial networks." arXiv preprint arXiv:1611.02163 (2016).
%\end{hangparas}
%\begin{hangparas}{0.2in}{1}
%\noindent [187] K. Basioti, and V. M. George. "Designing GANs: A likelihood ratio approach." arXiv preprint arXiv:2002.00865 (2020).
%\end{hangparas}
%\begin{hangparas}{0.2in}{1}
%\noindent [188] J. Engel, H. Matthew, and R. Adam. "Latent constraints: Learning to generate conditionally from unconditional generative models." arXiv preprint arXiv:1711.05772 (2017).
%\end{hangparas}
%\begin{hangparas}{0.2in}{1}
%\noindent [189] Z. Dai, A. Amjad, B. Philip, H. Eduard, and C. Aaron. "Calibrating energy-based generative adversarial networks." arXiv preprint arXiv:1702.01691 (2017).
%\end{hangparas}
%}

\end{document}